\documentclass{article} % To import the chapters from other file
\usepackage{graphicx}% Package to include the images
\usepackage{amsmath} % For math
\usepackage{amssymb}
\usepackage{commath} % for \abs
\usepackage{algorithm} % For algorithm
\usepackage{subcaption}
\usepackage{hyperref}
\usepackage{multirow}
\usepackage{pdflscape}
\usepackage{listings} % for  including codes from scripting languagues
\usepackage{multirow}
\usepackage[table,xcdraw]{xcolor}
\usepackage{algorithmic}
\usepackage{comment}
\usepackage{todonotes}
\DeclareMathOperator*{\argmax}{argmax}
\newcommand{\bx}{\ensuremath{\mathbf{x}}}

\newcommand{\bz}{\ensuremath{\mathbf{z}}}

\date{}

\title{Optimal Sepsis Patient Treatment using Human-in-the-loop Artificial Intelligence}

\usepackage[symbol]{footmisc}

\begin{document}
	\maketitle
	%\bigskip
\begin{center}	
Akash Gupta, PhD $^{1}$ \footnote[1]{Akash Gupta and Michael T. Lash equally contributed to this study.}, Michael T. Lash, PhD $^2$, Senthil K. Nachimuthu, MD, PhD, FAMIA$^3$

$^1$California State University, 18111 Nordhoff St, Northridge, CA 91330, USA, Email: akash.gupta@csun.edu

$^2$University of Kansas, 1645 Naismith Dr., Lawrence, KS 66045, USA, Email: michael.lash@ku.edu

$^3$3M Health Information Systems, Inc., 575 W Murray Blvd
Murray, UT 84123, USA, Email: snachimuthu@mmm.com

\end{center}
\begin{abstract}
    Sepsis is one of the leading causes of death in Intensive Care Units (ICU). The strategy for treating sepsis involves the infusion of intravenous (IV) fluids and administration of antibiotics. Determining the optimal quantity of IV fluids is a challenging problem due to the complexity of a patient's physiology. In this study, we develop a data-driven optimization solution that derives the optimal quantity of IV fluids for individual patients. The proposed method minimizes the probability of severe outcomes by controlling the prescribed quantity of IV fluids and utilizes human-in-the-loop artificial intelligence. We demonstrate the performance of our model on 1122 ICU patients with sepsis diagnosis extracted from the MIMIC-III dataset. The results show that, on average, our model can reduce mortality by 22\%. This study has the potential to help physicians synthesize optimal, patient-specific treatment strategies.
    
\end{abstract}

\section{Introduction}
		%\section{Background}
Sepsis is a life-threatening organ dysfunction caused by a dysregulated host response to infection. Each year, at least 1.7 million adults develop sepsis \cite{cdc}. Patients with sepsis are at considerable risk for severe complications and death; one in three hospital deaths are due to sepsis \cite{cdc}. The cornerstone of sepsis treatment is antibiotic administration and fluid resuscitation to correct hypotension. Studies have shown that the type of fluid resuscitation is correlated with mortality \cite{rochwerg2014}, and prescribing excess quantity of intravenous (IV) fluids to septic patients could be detrimental \cite{vincent2011}. Because of the significance of IV fluid in managing sepsis patients, it is critical to understand what \textit{type} of fluid and what  \textit{amount}  of fluid should be administered.  In this study, we develop a prescriptive clinical model that deduces optimal, patient-specific IV fluid values for the treatment of sepsis. %prediction-based optimization method, referred to as inverse classification, to derive the right fluid type and optimal quantity to be administered. 

The international community, Survival Sepsis Campaign (SSC), recommends early goal-directed therapy (EGDT) that underlines the importance of rapid volume resuscitation \cite{dellinger2013}. Among septic patients, IV therapy is employed as a volume expander in the event of blood loss to keep the body tissue oxygenated. There are two types of volume expanders: \textit{Crystalloids} and \textit{Colloids}. Since \textit{colloids} do not show clear benefits over crystalloids in treating critically ill patients despite their higher cost \cite{perel2012}, we aim at finding the right type of crystalloids. The types of crystalloids considered in this study are: Dextrose (a synonym for glucose) 10\% in water (D10W), Dextrose 5\% (D5W), Dextrose 5\% in Normal Saline (D5NS), 5\% Dextrose in half normal saline (D5HNS), Dextrose 5\% in Lactate Ringer (D5LR), Normal Saline (NS), Half Normal Saline (HNS), and Lactated Ringer (LR). Due to the complexity of patient physiology, there is no consensus on treatment strategy. The lack of treatment standards further complicates the treatment process for inexperienced junior doctors.

%Consider splitting into two paragraphs
In hospitals, junior doctors are usually responsible for initiating the immediate treatment of patients with severe sepsis for two reasons. First, on-call systems in hospitals are designed to first contact junior doctors when the condition of the patient deteriorates (often governed by the triggering of an early warning system). Second, junior doctors are often the first to attend to patients following hospitalization. SSC guidelines recommend varying the amount of fluid infusion at different disease severity levels. The study showed that 20\% of all adults receiving fluid resuscitation experience complications owing to inadequate quantity of fluid resuscitation \cite{courtney2014}. By the time senior doctors review the treatment, the early intervention window has passed, which is critical for sepsis treatment \cite{gupta2019}. Therefore, developing a clinical tool that can help junior doctors to deduce an optimal treatment is critical.

Some of the significant barriers to clinical models being accepted in the medical community are the issues of interpretability and explainability: medical practitioners expect a model to be transparent and should provide meaningful recommendations. Part of the novelty of our proposed method is that the model considers the physician's own treatment recommendation when eliciting an optimized recommendation. Furthermore, the model can be parameterized to only deviate a small amount from the physician-provided values. Thus, our physician in the loop formulation ensures that medical practitioners can trust the provided recommendations.

\begin{figure}[ht!]
    \centering
    \includegraphics[scale = 0.6]{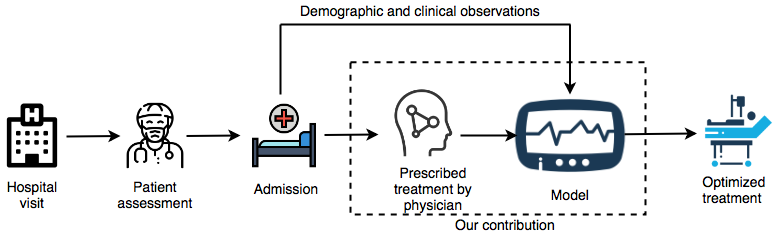}
    \caption{Patient flow in hospital}
    \label{fig:overview}
\end{figure}

The method is designed to provide recommendations that are tailored to each patient's physiology. Patient physiology is characterised using clinical data available in Electronic Health Records (EHR). EHR data is collected throughout a patient's hospital stay. Figure \ref{fig:overview} illustrates a patient's flow through a hospital system, and also highlights the contribution of this paper. Various clinical signs can be collected as the patient progresses through the hospital system. A patient with a suspected infection enters the hospital where the primary assessment of disease criticality is performed. Depending on the severity of the disease, a patient may be recommended for hospitalization. Alternatively, a patient who has been hospitalized may develop signs of infection during the inpatient stay. Following the SSC guidelines, the physician can recommend a specific amount of IV fluids. Our model combines the physician recommendation and the patient's clinical and demographic measurements to determine the optimal quantity of the IV fluids that improves survival probability.

%<Need to find out about Normal Saline and Sterile Water. Are they crystalloids ?>

This study makes the following contributions:
\begin{enumerate}
    \item Formulates a generalized optimization model to prescribe the optimal, patient-specific treatment.
    \item Develops a tool to derive the optimal quantity of IV fluids for septic patients in ICUs.
    \item Underlines the performance differences between the presence and the absence of human interactions in decision making using the proposed prescriptive model.
    \item Proposes a feature selection procedure that integrates well with the established modeling framework.
    \item Augments the theoretical literature on inverse classifiers.
\end{enumerate}

 Our devised human in the loop method is based on an \textit{inverse classification} framework \cite{lash2017a,lash2017b}. This framework includes two steps. First, some arbitrary classifier is trained that best predicts mortality. We employ logistic regression and deep learning classifiers in our experiments. Second, the selected model is embedded into an optimization formulation that takes patient data and physician-provided fluid resuscitation recommendations as input to derive the optimal IV fluid recommendations.  

The rest of this paper is structured as follows. Section \ref{sec:lit} reviews relevant literature, including existing IV fluid-based sepsis treatment recommendation models and applications of inverse classification to healthcare problems. Section \ref{sec:method} describes the proposed method in detail.  Section \ref{sec:dataprep} introduces the EHR data utilized in our experiments to evaluate our methodology. Section \ref{sec:results} presents the experimental settings and numerical results. Finally, Section \ref{sec:conclusion} summarizes the contribution of this study. 

\section{Literature Review} \label{sec:lit}

Due to complex pathophysiology, the treatment of sepsis imposes many challenges. As a result, the World Health Organization passed a resolution to improve the definition, diagnosis, and treatment of sepsis \cite{reinhart2017}. The SSC, in their guidelines, advocate for initial fluid resuscitation to treat sepsis effectively \cite{rhodes2017}. They specified that within three hours of sepsis detection, the patient should be prescribed 30 mL/kg of IV crystalloid fluid. However, the prescribed quantity should also be administered considering the patients' mass and other important vital signs \cite{cecconi2018}. Due to the threat sepsis imposes to human life, many studies have shown that the fluid amount should be considered critically rather than reflexively. Chang and Holcomb (2016) study different fluid resuscitation options available for treating sepsis patients \cite{chang2016}. The study discusses the basic physiology, science of intravenous fluid, and compares balance and unbalanced crystalloids. Durairaj and Schmidt (2008) indicated in their review paper that excessive fluid resuscitation can lead to adverse outcomes \cite{durairaj2008}. The authors recommended the use of a dynamic index to determine the fluid amount. This recommendation could be achieved by utilizing optimization models to prescribe patient-specific fluid amounts. 
 
Some researchers have applied machine learning techniques to determine optimal sepsis treatment strategies. Komorowski et al. (2018)  proposed a model to derive an optimal treatment strategy for sepsis using reinforcement learning \cite{komorowski2018}. The proposed model harvests the optimal treatment by comparing many treatment decisions. Raghu et al. (2017) extended (pre-print version) the work from \cite{komorowski2018} by employing the continuous state space solution approach to derive optimal sepsis treatment strategy \cite{raghu2017}. The authors optimize the quantity of IV fluids and vasopressors such that the treatment outcome can be improved. 
 
An optimization-driven tool could support junior doctors in determining optimal IV fluid treatments, who are usually among the first to treat patients. Courtney et al. (2014) studied the infusion of intravenous fluids to patients by junior doctors. The study found that the majority of junior doctors do not prescribe the right amount of fluid and fail to adjust for volume by mass \cite{courtney2014}. The results of this study highlight the importance of a convenient, human (physician) in the loop method that can aid junior physicians in making the right IV fluid prescription. 
 
A few studies have shown that early fluid resuscitation is associated with reduced mortality. Lee et al. (2014) studied the correlation between aggressive, optimal fluid resuscitation administered during the early stages of treatment \cite{lee2014}. The study found that earlier fluid resuscitation administered in adequate quantities is associated with decreased mortality. Cohen et al. (2015) suggested that, in the absence of any specific anti-sepsis treatment, the early administration of antibiotics and optimal infusion of fluid resuscitation is critical. This study designs a framework that can alert physicians to the risk of mortality during the early stages of a septic episode and assists physicians in prescribing an optimal treatment strategy using our devised inverse classification framework.

Recommender systems attempt to find the items, services, products, etc. that will lead an individual to be \textit{most satisfied}. Such methods are typically implemented in environments that contain an over-abundance of decisions, thereby leading users of the environment to a state of ``information overload''. For instance, the popular movie and TV streaming website Netflix employs a recommender system to filter through the thousands of entertainment options and display those that the customer will most enjoy. Notions of ``most satisified'', ``most enjoyed'', etc. are expressed in terms of some numerical measure. On Netflix, for instance, users express whether or not they enjoyed a particular TV show or movie by ``liking'' or ``disliking'' it. The recommender systems leverage a user's past ``likes'' and ``dislikes'' to recommend not-yet-watched movies and TV shows that align with the user's preferences. Broadly speaking, recommender systems literature can be decomposed into three categories: collaborative filtering methods \cite{su2009survey}, which make recommendations based on user information, content-based filtering methods \cite{lops2011content}, which make recommendations based on content information, and hybrid filtering methods \cite{burke2002hybrid}, which make recommendations based on both user and content information. 

In more recent years, recommender systems methodology has been incorporated into a deep learning paradigm. Deep learning methods can be either collaborative, content-based, or hybrid-based depending on the type of information used as input. These methods are best decomposed in terms of the type of neural network architecture employed. Deep recommender systems employing convolutions \cite{kim2016convolutional}, recurrent units \cite{ko2016collaborative}, ``vanilla'' MLPs \cite{chen2017locally}, and auto-encoders \cite{berg2017graph} have all been explored, as have graph-based methods \cite{zhang2019deep,berg2017graph}.

While recommender system methodology is important work to consider provided our problem setting, it is not best suited to such a setting since an over-abundance of decisions is not the issue. A related recommendation technology is referred to as inverse classification. Inverse classification makes use of an induced classification model to find the feature value perturbations that optimize for a particular outcome of interest (measured by the induced classifier); the feature value perturbations represent the recommendations. For instance, a model might be induced to learn the mapping from patient characteristics, such as age, blood pressure, and fluid amounts, to a disease outcome of interest, such as  survive/die from sepsis (as we do in this paper). Inverse classification will then use this model to find the perturbations to a new patient's feature values (i.e., IV fluids prescribed by the physician) that optimally minimize the probability of death due to sepsis.

Inverse classification has previously been applied to a variety of domains, including cardiovascular disease risk mitigation \cite{chi2012,yang2012,lash2017a,lash2017b, lash2018c}, hiring in nurseries \cite{aggarwal2010}, bankruptcy prediction and alleviation \cite{pendharkar2002}, diabetes \cite{barbella2009}, and student classroom performance \cite{lash2017a, lash2017b, lash2018c}. Several select works use inverse classification to explain rather than prescribe \cite{barbella2009,laugel2018comparison}. Methodologically speaking, past inverse classification works can be stratified by those employ constraints \cite{barbella2009,chi2012, lash2017a, lash2017b, mannino2000, yang2012} and those that do not \cite{aggarwal2010, pendharkar2002}. Constraints encourage solutions that are real-world feasible and personable (i.e., can be tailored to each individual's preferences) and are therefore desirable. Furthermore, some inverse classification methods are model-specific \cite{aggarwal2010, barbella2009, chi2012, mannino2000, pendharkar2002, yang2012}, relying on the use of a specific predictive model, while others are model agnostic \cite{lash2017a,lash2017b, lash2018c}. If a particular model is able to accurately map the probability space of a particular problem, then a model-specific inverse classification method may be most appropriate. The best/most accurate model is rarely known apriori and is therefore why model-agnostic inverse classification methods are generally preferred.

Several studies have shown that human interactions combined with machine learning methods can positively influence decision making \cite{holzinger2016}. Holzinger et al. (2016) performed  numerical experiments to show the advantages of human interactions in solving optimization problems and concluded that human-in-the-loop can significantly improve the solution  and the computation time \cite{holzinger2016b}. Duhaime (2016) advocates integrating humans and artificial intelligence to solve healthcare problems \cite{duhaime2016}. The author claims that such models are more reliable and trustworthy due to the blend of physicians' knowledge and artificial intelligence. 

\section{Proposed Methodology} \label{sec:method}
%\section{Optimal Patient-Specific Fluid Resuscitation Recommendations} \label{sec:optimizationModel}

In this section we disclose our proposed human in the loop method of eliciting optimal, patient-specific fluid resuscitation amounts to reduce the probability of death due to sepsis. The proposed prescriptive model first determines the best predictive model to estimate mortality probability. The selected model is then embedded in an optimization formulation to derive the optimal amount of IV fluid. We describe the selection of predictive model in Section \ref{sec:sep-out} and the optimization formulation in Section \ref{sec:optimizationModel}.  Prior to disclosing our framework we first provide some relevant preliminary notation.

%\subsection{Preliminaries}

Let $\{\bx^{(i)},y^{(i)}\}_{i=1}^{n}$ denote a dataset of $n$ instances (patient visits), where $\bx \in \mathcal{X} \subset \mathbb{R}^{p}$ and $y \in \{0,1\}$. A value of $y=1$ indicates that a patient has died of sepsis and $y=0$ indicates that a patient has survived. $p$ is the size of feature vector $\bx$, which represents the number of clinical variables. %The $p$ values composing each instance's $\bx$ feature vector represent patient-specific measurements. 
These include demographic measurements, such as age, vitals, such as blood pressure, lab-based measurements, such as serum creatinine, and the physician-prescribed treatment measurements.
%The aforementioned measurement categories (e.g., demographics) each have a unique bearing on sepsis patient outcomes (i.e.,~$y=0/1$). 
However, only specific feature categories can actually be manipulated to affect (i.e.,~change) the final outcome (survive or not). For instance, one cannot change his or her age, but we can change the amounts of the various IV fluids administered to the patient to improve their probability of survival. Additionally, while certain measurement categories cannot be directly manipulated, the values of the features in these categories may depend and vary according to feature values in other categories. For instance, blood pressure may be a function of both age and the administration of certain drugs.

We first ascribe notation to these different categories. Let $U$ denote the indices in $\bx$ that correspond to \textit{unchangeable} features, such as demographic information, $D$ denote the indices corresponding to the \textit{directly changeable} features, such as IV fluids, and $I$ denote \textit{indirectly changeable} features, such as vitals and lab-based measurements. Using these index sets the feature vector $\bx$ can be decomposed into $\bx_U,\bx_I,\bx_D$ to refer to specific feature values -- i.e.,~$\bx=\left(\bx_U,\bx_I,\bx_D\right)$. The feature sets of $\bx_U,\bx_I$ and $\bx_D$ are denoted using $\mathcal{F}_U$, $\mathcal{F}_I$ and $\mathcal{F}_D$, respectively; these notations are used predominantly in Section \ref{sec:featsel} to explain feature selection. 

\subsection{Predicting Sepsis Outcomes}\label{sec:sep-out}

With the above preliminaries explained, an initial model that provides probabilistic estimates of sepsis outcomes is formulated as follows:
\begin{align}
    \label{eq:fx}
    \hat{y} = f(\bx_U,\bx_I, \bx_D) = \frac{1}{1+e^{-g\left(\bx_U,\bx_I,\bx_D \right)}},
\end{align}
where $f(\cdot)$ is the logistic sigmoid function applied to another function $g(\cdot)$ that takes $\bx$ as input. Such a formulation allows for flexibility in defining the model used to make predictions, while still ensuring that probabilistic outputs are obtained from the model. A model that outputs probabilities is important since we wish to minimize the probability of a negative outcome directly, rather than minimizing over a set of discrete outcomes such as $\{\text{alive},\text{not alive}\}$.  When a logistic regression model is used, the $g(\cdot)$ function can be expressed as a linear combination of learned parameters and an instance feature vector:
\begin{align}
    \label{eq:g-logistic}
    g_{\text{logistic}}(\bx) = \left[b, \pmb{\theta} \right]^{\top}\left[ 1,\bx \right] = b+\sum_{j=1}^{p} \theta_j x_j, 
\end{align}
where $b$ and $\pmb{\theta}$ are the learned parameters. If $g(\cdot)$ is a more complex model, albeit still of the variety that can be trained to learn a probability mapping, like a neural network, the function can be expressed as:
\begin{align}
    \label{eq:g-nn}
    g_{\text{NN}}(\bx) = \mathbf{w}_k^{\top} h_{k-1} \left(\mathbf{W}_{k-1} \dots h_{1} \left(\mathbf{W}_{1} \bx \right) \right), 
\end{align}
where $\mathbf{w}_k$ is a parameter vector for the $k$th (i.e.,~output) layer, $\mathbf{W}_{j}:j=1,\dots,k-1$ are parameter matrices associated with the $k-1$ hidden layers, and $h_j:j=1,\dots,k-1$ are each some arbitrary, non-linear activation function (e.g., Rectified Linear Unit (ReLU), sigmoid, etc.). An activation function is a general non-linear mathematical function that transforms input in a way that is meant to mimic the firing of a neuron. The ReLu activation function, for instance, is defined as:

\begin{equation}
    h_k(\mathbf{W}_{k} \bx) = max(0, \mathbf{W}_{k} \bx) \nonumber
\end{equation}

When either $g_{\text{logistic}}$ or $g_{\text{NN}}$ are selected for use with $f$, then $f$ and $g$ are learned jointly through a gradient descent optimization process -- e.g., gradient descent, projected gradient descent, stochastic gradient descent (in the case of $g_{\text{NN}}$), etc. If the selected learning algorithm does not natively provide probability estimates, however (e.g., $g$ is an SVM model), then $g$ must be learned separately from $f$, and $f$ reduces to a specific application of Platt Scaling \cite{platt1999}.

Defining $f$ and $g$, as generally as we have, allows us to define a considerably larger hypothesis space $\mathcal{H}$ to search across when attempting to find the optimal model to use in our inverse classification framework. The model selection process can be briefly expressed as follows:
\begin{align}
    f^* = & \argmax_{f \in \mathcal{H}} \left \{ \text{Evaluate}\left( f, \mathbf{X}^{val}, c\right ) \right \} \label{eq:bestModel}
\end{align}

where the validation set $\mathbf{X}^{val} = \{\bx^{(i)}_{val}\}_{i=1}^{n_{val}}$ is evaluated on trained classifier, $f$, based on the classification measures, $c$. In our experiments, classification performance is measured using accuracy and the Area Under the Receiver Characteristic Curve (AUC), explained in Section \ref{sec:predictiveModel}. The increased search size of $\mathcal{H}$ is of significance since the quality of our life-saving recommendations are directly tied to the quality of the model and accuracy of the probability space mapping.%\footnote{As discussed in the related works section, a model-agnostic inverse classification framework is preferred for this reason}

\subsection{Inverse Classification for Optimal Dose Prescriptions} \label{sec:optimizationModel}

Using the general model defined in Equation \eqref{eq:fx} we can obtain probabilistic estimates of death due to sepsis $\hat{y}$ for some instance $\bx$. This model will form the basis for our method of eliciting optimal, patient-specific fluid resuscitation recommendations, which we discuss in this subsection.

Our recommendation formulation is based on inverse classification, which is the process of manipulating the feature values of an instance in order to minimize the probability of an undesirable outcome. In this case, the instances are ``sepsis patients'' and the undesirable outcome is ``death''. Therefore, our formulation will minimize the probability of death, which can be expressed:
\begin{align}
    \label{eq:init-ic}
     \min_{\tilde{\bx}} \hspace{.2cm} & f\left( \tilde{\bx}\right)
\end{align}
where $\tilde{\bx}$ is a test instance whose feature values are freely manipulable by the optimization process.%\footnote{We discuss our optimization methodology and assumptions later on in Section \ref{sec:rw-frame}.}

The initial formulation, Equation \eqref{eq:init-ic}, does not include real-world feasibility considerations, however. For instance, it would make little sense to allow the formulation to manipulate the $\bx_{U}$ feature values, such as age. Moreover, the extent of changes allowed by Equation \eqref{eq:init-ic} is unbounded and may produce nonsensical recommendations, such as negative fluid amounts, as a result. Finally, Equation \eqref{eq:init-ic} doesn't take into account variable dependencies and interactions. As an example, blood pressure, an indirectly changeable feature, is a function of the unchangeable features $\bx_U$, such as age, and the directly changeable features $\bx_D$, representing IV fluids. Since we are manipulating the $\bx_D$ (i.e. IV fluids), we can expect blood pressure to also change as a result, and because one's blood pressure has an impact on whether one lives or dies, it is critical to capture these dependencies during optimization

%\subsubsection{Indirect Feature Estimator}

To account for the dependencies that exist between $\bx_U,\bx_D$ and $\bx_I$, as discussed in the preceding paragraph, we propose to use a so-called \textit{indirect feature estimator} (IFE). Let $H:\mathcal{X}^{U,D} \rightarrow \mathcal{X}^{I}$ denote a function that takes the $U$ and $D$ features as input and provides estimates for the $I$ features, i.e.,

\begin{equation}
    \hat{\bx}_{I}=H\left(\bx_U,\bx_D \right). \label{eq:h_function}
\end{equation}

Using $H(\cdot)$, we can account for how changes to the $\bx_D$ features affect the $\bx_I$ features. The IFE can be any differentiable regression model and is therefore also fairly flexible. To be more concrete, we explicitly require a differentiable model because our optimization methodology relies on gradient information. In this study, we employ linear regression and neural networks. If we need to relax the assumption of differentiablility then the optimization methodology would need to be adjusted accordingly (e.g.,~heuristic optimization). Relaxing this assupmtion may broaden the IFE hypothesis space at the cost of optimality guarantees.

%\subsubsection{A Framework for Real-world Feasible Recommendations \label{sec:rw-frame}}

Considering that 1) the $\bx_I$ feature values are governed by $\bx_U$ and $\bx_D$ and 2) only the $\bx_D$ feature values can be manipulated, the naive inverse classification formulation of \eqref{eq:init-ic} can be transformed into:
\begin{align}
    \label{eq:ic2}
     \min_{\tilde{\bx}_D} \hspace{.2cm} & f\left( \bx_U, H \left(\bx_U,\tilde{\bx}_D\right), \tilde{\bx}_D\right),
\end{align}
where $\tilde{\bx}_D$ are now the decision variables (to reflect the fact that only these variables should be changed).

While Equation \eqref{eq:ic2} is an improvement over Equation \eqref{eq:init-ic}, there are still necessary considerations missing from the formulation. Namely, Equation \eqref{eq:ic2} is unconstrained, which may allow the formulation to produce nonsensical recommendations, such as negative fluid resuscitation amounts. Additionally, since a key component of our method is human in the loop functionality (i.e.,~``doctor in the loop'' functionality), we wish to restrict the \textit{extent} of the manipulations made to the expert-specified $\bx_D$ feature values. To be a bit more explicit, our rationale is that a doctor's prescribed fluid resuscitation amounts reflect a coarse-grained recommendation. Our inverse classification method will then refine this initial, doctor-specified recommendation to provide a fine-grained, precise recommendation that does not deviate ``too far'' from that specified by the doctor.

Therefore, we update Equation \eqref{eq:ic2} by adding feasibility constraints to prevent nonsensical recommendations as follows:
\begin{align}
    \label{eq:opt-ic}
    \min_{\tilde{\bx}_{D}} \hspace{.2cm} & f\left(\bx_{U}, H\left(\bx_U,\tilde{\bx}_{D} \right), \tilde{\bx}_{D} \right) \\ \nonumber
    \text{s.t.} \hspace{.2cm} & \Vert \bz \Vert_{1} \leq b \\
    \hspace{.2cm} & \mathbf{0} \leq \tilde{\bx}_D \leq \mathbf{1} \nonumber
\end{align}
where $\bz = \tilde{\bx}_{D} - \bx_D $ and $b$ is a budget term that controls the extent of recommendations allowed. Note that $\tilde{\bx}_{D}$ are the updated feature values and $\bx_D$ are the physician-provided feature values. 

The $b$ constraint is best calibrated to each physician user in an offline setting, prior to deployment. If a physician is more experienced, the $b$ term can likely be smaller than if the physician is less experienced. A smaller $b$ term will produce smaller recommendations since the cumulative change recommended is restricted. 

\begin{figure}[!htp]
    \centering
    \includegraphics[scale=0.50]{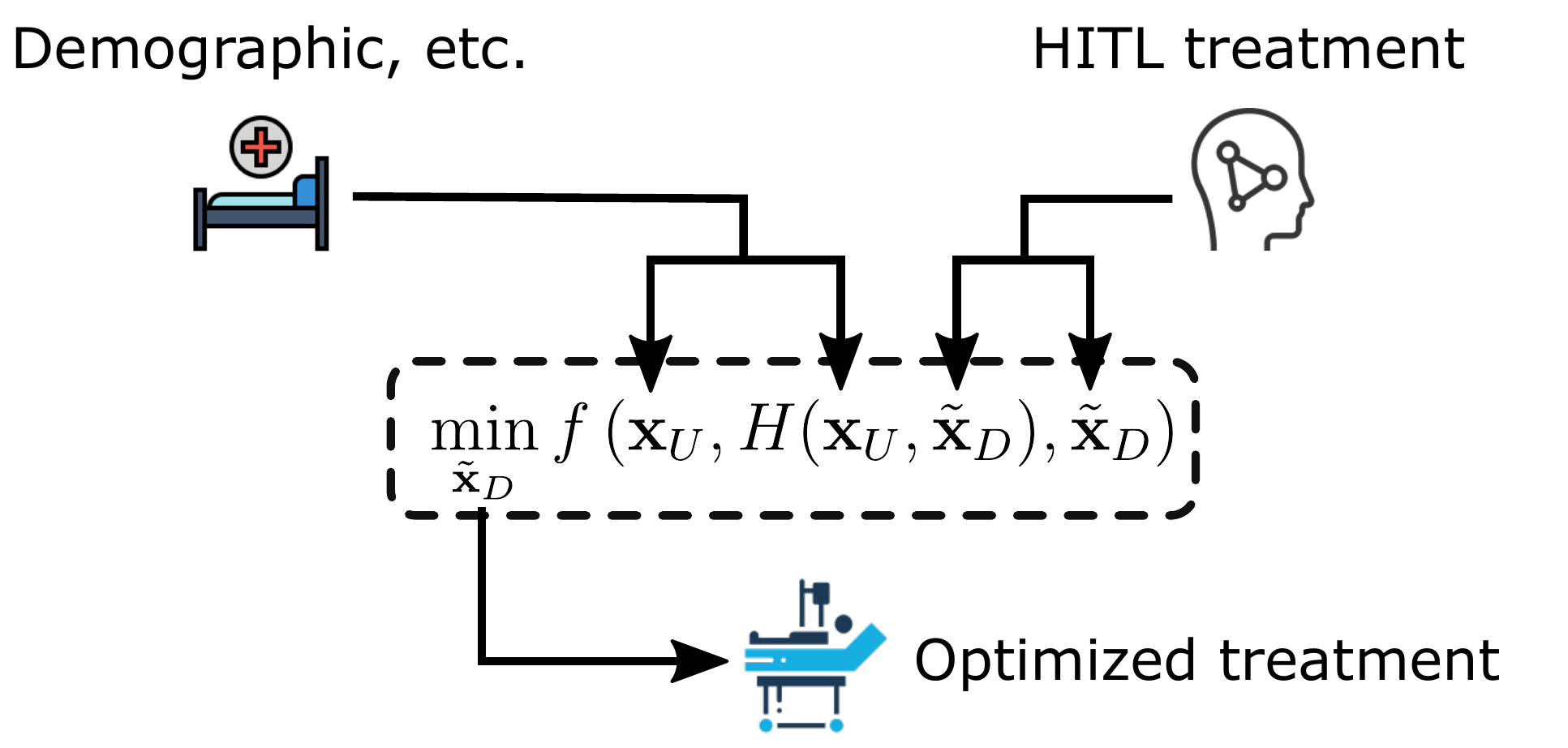}
    \caption{A depiction of our human in the loop formulation. HITL: human in the loop}
    \label{fig:ic-proc}
\end{figure}

The last line of Equation \eqref{eq:opt-ic} specifies that recommendations must be non-negative, but also less than or equal to one. This latter consideration is based on the assumption that all features have been normalized to a zero-one range and allows us to further encourage the process to produce real-world feasible recommendations. Figure \ref{fig:ic-proc} illustrates the inverse classification formulation in terms of the feature value categories. Note that the human in the loop (HITL) treatments denote the physician-provided IV fluid values.

The final  formulation, Equation \eqref{eq:opt-ic}, can be optimized using project gradient descent (PGD), which is an efficient gradient-based optimization method \cite{Nesterov07composite}. Therefore, the updates to the $\tilde{\bx}_D$ feature values at each iteration $t=1,\dots,T$ of PGD can be expressed by:
\begin{align}
    \label{eq:opt-pgd}
    \tilde{\bx}_{D}^{(t+1)} &= \Pi_{\Delta} \left( \tilde{\bx}_{D}^{(t)} - \eta^{(t)} \frac{\partial f}{ \tilde{\bx}_{D}^{(t)}} \right) \\ \nonumber
    & = \Pi_{\Delta}  \left(\tilde{\bx}_{D}^{(t)} - \eta^{(t)} \left[ \frac{\partial f}{\partial  \tilde{\bx}_{D}^{(t)}} + \frac{\partial f}{\tilde{\bx}_I^{(t)}} \frac{\partial H}{ \tilde{\bx}_{D}^{(t)}} \right] \right)\\
    & = \Pi_{\Delta}  \left( \tilde{\bx}_{D}^{(t)} - \eta^{(t)} \nabla f \right) \nonumber
\end{align}
where $\Pi_{\Delta}  \left( \cdot \right)$ is the projection operator that projects the input onto the feasible region $\Delta$, $\tilde{\bx}_I^{(t)} = H\left(\bx_U,\tilde{\bx}_D^{(t)} \right)$, and $\eta$ is a learning rate. Note that the gradient depends in part upon the estimate of the $\bx_I$ feature values. Further note that the projection done by $\Pi\left(\cdot \right)$ onto the feasible region $\Delta$, dictated by the constraints in \eqref{eq:opt-pgd}, can be achieved efficiently and always succeed (as long as the feasible region $\Delta \neq \varnothing$) \cite{lash2017b}. %by setting each $c_i = 1$.

\section{Data Preparation \label{sec:dataprep}}
\subsection{Data}

This study extracted EHR data from MIMIC III (Medical Information Mart for Intensive Care III) \cite{johnson2016} to demonstrate the performance of the proposed methodology. These clinical data are freely-available de-identified electronics health records consisting of over sixty thousand patients (61,532) hospitalized in critical care units (ICU) at the Beth Israel Deaconess Medical Center between 2001 and 2012. We selected the adult patient visits with a diagnosis of sepsis for use in our study.

\subsection{Data pre-processing}	

\begin{figure}[ht!]
	    \centering
	    \includegraphics[scale = 0.6]{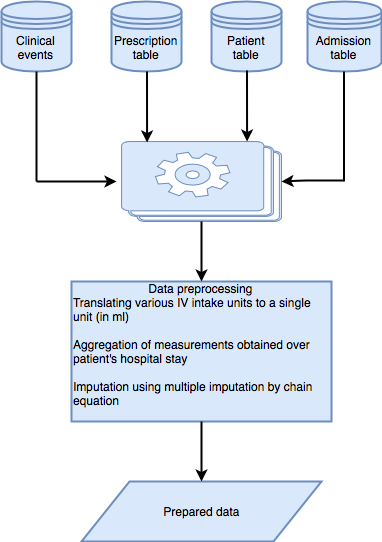}
	    \caption{Data preparation steps}
	    \label{fig:dataPrepSteps}
	\end{figure}

 Several necessary pre-processing steps were taken to curate our final dataset. Figure \ref{fig:dataPrepSteps} illustrates these data preparation steps. The data, initially available in multiple tables, are merged to consolidate information about patients, clinical observations and treatments. A particular challenge when pre-processing medical data is the plethora of terms conveying the same meaning (synonyms), which must be accounted for when deriving a suitable dataset for this study. For example, respiratory rate is recorded as \textit{Respiratory Rate, Breath rate, Res. rate etc.}.
 
 Ultimately, the derivation of the dataset used in our study is performed using a clinical expert's opinions and by consulting the relevant literature \cite{komorowski2018, gupta2020}. For each patient visit, the clinical variables are recorded longitudinally (i.e., are measured across time). We then aggregate the longitudinal data by computing the mean of each clinical variable for each visit. This aggregation approach was also performed for IV fluids. Any missing data is imputed using multiple imputation by using the chain equation approach \cite{buuren2010}. 

Pre-processing the data results in a total of 1122 patient visits with a diagnosis of sepsis. We extracted information about 30 features for these patient visits. The statistical summary (minimum, first quarter, second quarter or median, mean, third quarter and maximum) of each feature is included in Table \ref{tab:statisticalSummary}. The mean and the median age of patients is 68 and 79 years, respectively. Among all patient visits, the outcome for 244 (22\%) visits is expired. The data include a large proportion of patient visits with severe outcome because the study focuses on ICU patients.

\begin{table}[ht!]
\centering
\caption{Statistical summary of clinical features (Q1: first quarter, Q2: second quarter, Q3: third quarter)}
\label{tab:statisticalSummary}
\begin{tabular}{lrrrrrr}
\hline
Clinical Variables (units)          & Minimum & Q1    & Median  or Q2 & Mean  & Q3    & Maximum \\
\hline
Base Excess (mEq/L)                 & -31.0   & -5.0  & -2.3          & -2.7  & 0.0   & 16.5    \\
Blood CO\textsubscript{2} (mEq/L)                         & 4.5     & 19.8  & 22.6          & 22.8  & 25.8  & 44.0    \\
Blood Hemoglobin (g/dL)                  & 6.0     & 9.1   & 9.9           & 10.1  & 11.0  & 19.5    \\
Blood Urea Nitrogen (mg/dL) & 1.0     & 17.0  & 29          & 36.01  & 48.2  & 212.5   \\
Body Temperature ($^{\circ}$F) & 47.4    & 97.5  & 98.1          & 98.1  & 98.8  & 107.2   \\

Diastolic Blood Pressure (mmHg)    & 18.0    & 50.8  & 56.4          & 57.1  & 63.0  & 90.3    \\
Glasgow Coma Scale Score    & 3.0     & 10.6  & 13.9          & 12.5  & 15.0  & 15.0    \\
Heart Rate (/min)                  & 46.6    & 78.4  & 88.2          & 89.0  & 98.9  & 137.3   \\
Hematocrit (\%)                  & 19.7    & 27.8  & 29.9          & 30.7  & 32.9  & 61.6    \\
Lactate (mg/dL)                    & 0.6     & 1.5   & 2.0           & 2.5   & 2.9   & 18.3    \\
O\textsubscript{2} Flow (L/min)                    & 0.3     & 2.2   & 3.4           & 5.1   & 6.3   & 100.0   \\
PaCO\textsubscript{2} (mmHg)                      & 19.0    & 33.6  & 38.5          & 39.8  & 43.8  & 121.0   \\
PaO\textsubscript{2} (mmHg)                       & 27.0    & 83.0  & 104.1         & 108.1 & 127.5 & 350.0   \\

pH                          & 2.4     & 7.3   & 7.4           & 7.4   & 7.4   & 53.5    \\
PO\textsubscript{2} (mmHg)                         & 26.0    & 73.3  & 100.0         & 103.9 & 126.7 & 467.0   \\
PT (s)                          & 11.6    & 13.8  & 14.9          & 16.6  & 17.6  & 55.2    \\
PTT (s)                         & 14.9    & 24.0  & 27.9          & 31.7  & 35.4  & 128.3   \\
Platelet count (x1000/mm\textsuperscript{3})                   & 16.8    & 145.7 & 215.3         & 227.5 & 288.0 & 985.0   \\
Respiratory Rate (/min)            & 10.7    & 17.7  & 20.3          & 20.5  & 22.9  & 38.1    \\
Serum Creatinine (mg/dL)                  & 0.2     & 0.8   & 1.2           & 1.9   & 2.0   & 141.9   \\
Serum Chloride (mEq/L)                    & 84.0    & 102.6 & 106.2         & 106.1 & 109.6 & 137.6   \\
Serum Glucose (mg/dL)                     & 30.3    & 107.9 & 126.7         & 135.4 & 150.8 & 447.7   \\
Serum Magnesium (mEq/L)                   & 1.1     & 1.8   & 2.0           & 2.0   & 2.1   & 18.3    \\
Serum Potassium (mEq/L)                   & 2.7     & 3.7   & 4.0           & 4.1   & 4.3   & 7.3     \\
Serum Sodium (mEq/L)                      & 118.3   & 136.9 & 139.2         & 139.3 & 141.8 & 163.1   \\
Systolic Blood Pressure (mmHg)     & 0.0     & 102.2 & 109.9         & 111.7 & 120.7 & 210.1   \\

WBC Count (x1000/mm\textsuperscript{3})                         & 0.5     & 8.3   & 11.7          & 13.2  & 15.8  & 97.1    \\

%Urea Nitrogen               & 3.2     & 16.6  & 27.1          & 32.2  & 42.3  & 165.0   \\

Age (years)              & 19.0    & 56.3  & 68.0          & 66.4  & 79.0  & 89.0  \\
Median Weight (kg)          & 0.0     & 63.8  & 76.8          & 80.3  & 90.8  & 233.9   \\

\hline
\end{tabular}
\end{table}

Besides demographics, vitals, and lab-based data, we also extracted the physician-prescribed dosage of IV fluids. Table \ref{tab:iv_fluid} lists the names of IV fluids, as recorded in the dataset. As mentioned earlier, a key medical data challenge is the use of synonymous terms; many IV fluids are recorded with different names, but refer to the same fluid. We translate these synonymous fluid names into a standard convention following a clinical expert's suggestions. This study incorporates nine different types of IV fluids. Additives/solutes (e.g. potassium chloride) that were administered via an IV infusion were parsed to obtain the type of the underlying IV fluid. We assume that the absence of a specific type of IV fluid for a patient indicates that the specific type of IV fluid was not prescribed. 
%{\color{blue} include information about IV fluids}
%\subsection{Variables}
\begin{table}[ht!]
\caption{Available types of IV fluids in the dataset}
\label{tab:iv_fluid}
\begin{tabular}{lrr}
\hline
IV name (in dataset)                     & Standardized IV fluid name      & Standardized acronym \\
\hline
D10W                                     & Dextrose 10\% in Water          & D10W                 \\
D5 1/2NS                                 & D5 1/2NS                        & D5HNS                \\
Potassium Chl 20 mEq / 1000 mL D5 1/2 NS & D5 1/2NS                        & D5HNS                \\
Potassium Chl 40 mEq / 1000 mL D5 1/2 NS & D5 1/2NS                        & D5HNS                \\
D5LR                                     & Dextrose 5\% in Lactated Ringer & D5LR                 \\
D5NS                                     & Dextrose 5\% in Normal Saline   & D5NS                 \\
Iso-Osmotic Dextrose                     & Dextrose 5\%                    & D5W                  \\
D5W                                      & Dextrose 5\%                    & D5W                  \\
5\% Dextrose                             & Dextrose 5\%                    & D5W                  \\
Dextrose 5\%                             & Dextrose 5\%                    & D5W                  \\
D5W (EXCEL BAG)                          & Dextrose 5\%                    & D5W                  \\
5\% Dextrose (EXCEL BAG)                 & Dextrose 5\%                    & D5W                  \\
Potassium Chl 40 mEq / 1000 mL D5W       & Dextrose 5\%                    & D5W                  \\
Amino Acids 4.25\% W/ Dextrose 5\%       &                                 & DNS                  \\
1/2 NS                                   & Half normal saline              & HNS                  \\
0.45\% Sodium Chloride                   & Half normal saline              & HNS                  \\
LR                                       & Lactated Ringers                & LR                   \\
Lactated Ringers                         & Lactated Ringers                & LR                   \\
NS                                       & Normal Saline                   & NS                   \\
SW                                       & Normal Saline                   & NS                   \\
0.9\% Sodium Chloride                    & Normal Saline                   & NS                   \\
0.9\% Sodium Chloride (Mini Bag Plus)    & Normal Saline                   & NS                   \\
NS (Mini Bag Plus)                       & Normal Saline                   & NS                   \\
Iso-Osmotic Sodium Chloride              & Normal Saline                   & NS                   \\
Isotonic Sodium Chloride                 & Normal saline                   & NS                   \\
NS (Glass Bottle)                        & Normal Saline                   & NS                   \\
Potassium Chl 40 mEq / 1000 mL NS        & Normal Saline                   & NS                   \\
Potassium Chl 20 mEq / 1000 mL NS        & Normal Saline                   & NS                  \\
\hline
\end{tabular}
\end{table}

\subsection{Feature Selection \label{sec:featsel}}

Oftentimes a model with superior predictive performance can be produced by including/excluding certain features. The process of discovering the group of features that produces this superior model is called \textit{feature selection}. Because the inverse classification procedure relies on an underlying model to make life-saving fluid resuscitation recommendations, it is of paramount importance that the most accurate model be learned. Therefore, we propose to employ feature selection methodology to further improve model performance and thereby the accuracy of the probability space mapping captured by the learned model. Therefore, we propose a \textit{Classifier Subset Evaluation}-based (CSE) feature selection method, disclosed by Algorithm \ref{algo:cse}. Our CSE method can be viewed as a specific implementation of the more general \textit{ClassifierSubsetEval} method found in Weka \cite{weka}. %If one uses forward selection, a greedy stepwise optimization scheme, and sets the use of a validation set to false, then the implementation in \cite{weka} is equivalent to our method.

Algorithm \ref{algo:cse} takes as input a training set $\{ \bx^{(i)}, y^{(i)}\}_{i=1}^{n}$, the full feature set $\mathcal{F}$, an arbitrary classifier $f_{\pmb{\theta}}$ with any necessary, user-specified parameters $\pmb{\theta}$, a classifier performance measure $c$, such as accuracy or AUC, and $s$, which specifies the number of consecutive non-improving iterations allowed before termination. Upon execution, the algorithm initializes $\mathcal{F}_{opt}$, which will hold the selected features, to an empty set, $k$, which counts the number of iterations, is initialized to zero, the termination criteria $term$ is initialized to false, $c^{(k)}$ is the performance of the classifier trained at the $k$th iteration with $c^{(0)}$ initialized to 0, $\delta$, which measures classifier improvement in terms of $c$ by adding the feature at the $k$th iteration (i.e., $c^{(k)} - c^{(k-1)}$), is initialized to infinity,and $s^{\prime}$, which represents the number of successive, non-improving iterations currently observed, is set to 0. From here the algorithm begins iterating until $term$ is set equal to true, the conditions for which are expressed on lines 14 and 15 and will be explained momentarily.

\begin{algorithm}[!htp]
	\caption{Classifier Subset Evaluation $\textbf{ClassSubEval}\left( \{\bx^{(i)}, y^{(i)}\}_{i=1}^{n}, \mathcal{F}, f_{\pmb{\theta}}, c, s \right)$}
	\label{algo:cse}
	\begin{algorithmic}[1]
		\REQUIRE $\{ \bx^{(i)}, y^{(i)}\}_{i=1}^{n}$, $\mathcal{F}$, $f_{\pmb{\theta}}$, $c$, $s$
		\STATE Initialize $\mathcal{F}_{opt} \leftarrow \{\}$, $k \leftarrow 0$, $\delta \leftarrow \infty$, $term \leftarrow False$, $c^{(0)} \leftarrow 0$, $s^{\prime} \leftarrow 0$
		\WHILE{$term == False$}
		    \STATE $k \leftarrow k+1$, $\mathtt{f} \sim \mathcal{U}\left(\mathcal{F} \right)$, $\mathcal{F} \leftarrow \mathcal{F} \setminus \mathtt{f}$
		    \STATE $\mathcal{F}_{opt} \leftarrow \mathcal{F}_{opt} \cup \mathtt{f}$
		    \STATE $c^{(i)} \leftarrow train^{c} \left(\{ \bx^{(i)}, y^{(i)}\}_{i=1}^{n}, f_{\pmb{\theta}}, \mathcal{F}_{opt} \right)$
		    \STATE $\delta \leftarrow c^{(i)} - c^{(i-1)}$
		    \IF {$\delta \leq 0$ \AND $\vert \mathcal{F}_{opt} \vert \neq 1$}
		        \STATE $\mathcal{F}_{opt} \leftarrow \mathcal{F}_{opt} \setminus \mathtt{f}$
		        \STATE $\mathcal{F} \leftarrow \mathcal{F} \cup \mathtt{f}$
		        \STATE $s^{\prime} \leftarrow s^{\prime} + 1$
		    \ELSE
		        \STATE $s^{\prime} \leftarrow 0$
		    \ENDIF
		    \IF{$s^{\prime} == s$ \OR $\mathcal{F} == \varnothing$}
		        \STATE $term \leftarrow True$
		    \ENDIF
		\ENDWHILE
		\ENSURE $\mathcal{F}_{opt}$
	\end{algorithmic}
\end{algorithm}

At each iteration, $k$ is incremented by one, a feature $\mathtt{f} \sim \mathcal{U}\left(\mathcal{F} \right)$ is randomly selected from $\mathcal{F}$, and $\mathcal{F}$ is updated to exclude the selected feature ($\mathcal{F} \leftarrow \mathcal{F} \setminus \mathtt{f}$) (line 3). Next (line 4), the selected feature is added to $\mathcal{F}_{opt}$ where, subsequently, a model $f_{\pmb{\theta}}$ is trained using only the $\mathcal{F}_{opt}$ features (line 5) and evaluated in terms of metric $c$ to obtain $c^{(i)}$; $\delta$ is then computed on line 6. On line 7, $\delta$ is evaluated to see if adding $\mathtt{f}$ has improved predictive performance (this occurs when $\delta$ $>$ 0). If $\mathtt{f}$ does not improve the model ($\delta$ = 0), or worsens model performance ($\delta$ $<$ 0), then the addition of $\mathtt{f}$ to $\mathcal{F}_{opt}$ is undone: $\mathtt{f}$ is re-added to $\mathcal{F}$ (lines 8-9), and $s^{\prime}$ is incremented (line 10); otherwise (i.e., $\delta$ $>$ 0) $s^{\prime}$ is set to zero (lines 11-12). Line 14 specifies the termination criteria: if $s$ concurrent iterations have failed to produce an improvement or if all features have been added to $\mathcal{F}_{opt}$. Utilizing the outlined CSE-based feature selection method, we improve the predictive performance of our model and thereby the accuracy of the probability space mapping. %In the next section, which includes our results, we experiment with using our feature selection method.

\section{Results \label{sec:results}}

Previously, we elaborated on our proposed methodology and the selected clinical dataset. In this section we illustrate the performance of our model on this clinical dataset. The experimental results are stratified into five segments. First, we experiment with our CFS-based feature selection method while searching for the best predictive model. The best model and the selected subset of features are used in subsequent experiments. Second, we conduct experiments to find the best IFE. Third, using our optimal predictive model and optimal IFE, we derive optimized treatments relative to budget constraint $b$. %we evaluate our human in the loop (HITL) method in terms of probability improvement relative to budget term $b$. 
Fourth, we perform a robustness check of our model to evaluate performance should physician recommendations not be available.
Finally, we present the average changes recommended by our method relative to several selected budget constraint values $b$.

All results were obtained by first randomly partitioning our dataset into training, validation, and test sets. Since only approximately 20\% of the dataset instances (patient visits) belong to the positive class, we ensured that equal proportions of positive instances were allocated to each set (i.e., 20\% of the instances in each training, validation, testing set are positive). Dataset sizes were selected to be 80\%, 10\%, and 10\% for training, validation, and test sets, respectively. All $f$ and $H$ models were trained using the training set. The best type of each model ($f^*$, $H^*$) was selected based on validation set performance. The testing set was reserved for evaluating our recommendations and was not used in constructing or selecting predictive models. Finally, all features are normalized to a range of $[0,1]$ using min-max scaling.

\subsection{Variable Selection}\label{sec:featSel}

As discussed in Section \ref{sec:sep-out}, we need to search across $\mathcal{H}$ and select the best predictive model ($f^*$) to estimate the probability of mortality (Equation \ref{eq:bestModel}). Our first model-building step is to eliminate clinical features that do not contribute to the prediction, and may even detract from predictive performance. Therefore, we establish a variable selection procedure (outlined in a preceding section) that integrates well with our problem setting (and is one of the contributions of this paper).

Table \ref{tab:varList} lists both independent (or predictors) and dependent (or response) features. The independent variables are divided into three categories: $\bx_D$, $\bx_I$ and $\bx_U$. In our dataset, the amount and the type of the prescribed IV fluid are under direct control of the physician; hence, such variables are in the category of directly changeable features ($\bx_D$). The vitals (e.g., heart rate, blood pressure) and lab-based measurements (e.g., creatinine, white blood cell count) can not be manipulated directly, but can be manipulated \textit{indirectly} through manipulation to the administered IV fluids (the $\bx_D$ features). Therefore, all patient vitals and lab-based measurements fall under the category of indirectly changeable features ($\bx_I$). The dataset also includes variables such as age, gender etc. These patient attributes can not be altered. Hence, demographic features fall under the category of unchangeable features ($\bx_U$). 

Feature selection was performed using CSE, discussed in Section \ref{sec:featsel}. CSE can be applied to each $f \in \mathcal{H}$ in an ``online'' fashion, or as a pre-processing step, where a single $f$ is selected and used to find the subset of features that will be used when searching for $f^* \in \mathcal{H}$. We adopt the latter, pre-processing strategy to reduce experiment compute time. Our selected $f$ was a neural network model trained for 150 epochs with a single hidden layer containing three hidden nodes. This particular parameterization of $f$ was selected because it is representative of the parameterizations explored during the model tuning phase, the results of which are discussed in the next subsection.

Table \ref{tab:varList} shows the variables before and after employing the variable selection procedure. The variable selection results show that \textit{D5HNS, D5LR, D5W, LR} and \textit{NS} are the only fluids that affect the probability of mortality significantly. We also observe that the size of the indirect variable set has been reduced from 27 to 20 features. \textit{Discharge type} is considered the outcome, or dependent feature. In the next subsection, we show the performance of the predictive model on the data with the complete set and on the data with the selected features. 

\begin{table}[ht!]
		\centering
		\caption{Clinical features in the dataset}
		\label{tab:varList}
		\begin{tabular}{llp{5.2cm}p{5.2cm}}
			\hline 
			Variable category & Notation & Complete set of variables & Selected variables\\
			\hline
			Independent  &$\bx_D$ & Amount of fluid resuscitation (D10W, D5HNS, D5LR, D5NS, D5W, DNS, HNS, LR and NS) & D5HNS, D5LR, D5W, LR, NS\\
			&$\bx_I$  & Base excess, blood CO\textsubscript{2}, blood hemoglobin, blood urea nitrogen, body temperature, diastolic blood pressure, Glasgow Coma Scale, heart rate, hematocrit, lactate, O\textsubscript{2} flow, PaCO\textsubscript{2}, PaO\textsubscript{2},  pH, PO\textsubscript{2}, PT, PTT, 
			platelet, respiratory rate, serum creatinine, serum chloride, serum glucose, serum magnesium,   serum potassium, serum sodium, systolic blood pressure,   WBC       (27 variables) &
			
			Base excess, blood CO\textsubscript{2}, blood hemoglobin, blood urea nitrogen, diastolic blood pressure, Glasgow Coma Scale, heart rate, hematocrit, lactate, PaCO\textsubscript{2}, PaO\textsubscript{2}, PO\textsubscript{2}, PTT, 
			platelet, respiratory rate, serum creatinine, serum glucose, serum sodium, systolic blood pressure,   WBC  (20 variables) \\
			
			%Heart rate, systolic blood pressure, diastolic blood pressure, Glasgow Coma Scale, respiratory rate, creatinine, white blood cell count, sodium, glucose, CO2, hemoglobin, hematocrit, platelet, PTT, lactate, urea nitrogen, base excess, PaO2, PO2, PaCO2 (21 variables)\\
			&$\bx_U$&Age, gender, weight & Age, gender\\
			Dependent  & $y$& Discharge type (binary)& Discharge type\\
			\hline 
		\end{tabular}
\end{table}

\subsection{Predictive Model Tuning} \label{sec:predictiveModel}

In Section \ref{sec:featSel}, CFS is utilized to find the best subset of features. In this section, we employ a grid search to find the model $f^*$ that has the best predictive performance and therefore the best probability space mapping. We apply the grid search to two datasets: a dataset containing the full set of features and a dataset containing only those features that were selected using CFS. This will allow us to choose not only the best model, but to assess whether CFS is in fact able to produce a superior model. We limit our study to logistic regression ($g_{\text{logistic}}(\bx)$) and neural network variants ($g_{\text{NN}}(\bx)$). After determining the optimal $f^*$ and dataset, a grid search is performed to find the optimal IFE function $H^*$. We limit this grid search to multivariate linear regression and variants of neural networks.

Accuracy and AUC are employed to assess the performance of our classification models $f \in \mathcal{H}$. Accuracy is defined as the ratio obtained by dividing the number of  correctly predicted instances by the total number of instances. As a metric, however, accuracy is susceptible to class imbalance, which is present in our dataset (i.e., only 20\% of instances are positive). Therefore, we also adopt the AUC metric, which is insensitive to class imbalance. AUC plots the true positive rate (TPR) against the false positive rate (FPR), thus creating a curve. The area underneath this curve (called the receiver operating characterstic curve) is the AUC (area under the curve). When a model predicts \textit{only} the majority class (e.g., the model always predicts the negative class), the AUC is 0.50 and is why AUC is considered insensitive to class imbalance. An AUC of 1.0 represents completely perfect predictions.

\begin{table}[ht!]
\centering
\caption{Grid search results for both the original and ``feature selection'' datasets on a randomly held out validation set. %\textcolor{red}{Red} denotes the best \textcolor{red}{AUROC} for each dataset and \textcolor{blue}{blue} denotes the best \textcolor{blue}{accuracy}. 
HN: Hidden nodes, Tr Acc: Accuracy on training data, Tr AUC: AUC on training dataset, Val Acc: Accuracy on validation dataset, Val AUC: AUC on validation dataset}
\label{tab:modres}
\begin{tabular}{p{0.4in}|rr|rrrr|rrrr}\hline
 Models& \multicolumn{2}{c|}{Parameters} & \multicolumn{4}{c|}{All Features} & \multicolumn{4}{c}{Selected Features} \\ \hline
&HN & Epochs & Tr Acc & Tr AUC & Val Acc & Val AUC & Tr Acc & Tr AUC & Val Acc & Val AUC \\ \hline
\multirow{4}{*}{Logistic} & & 100 & 0.8185 & 0.8988 & 0.7768 & 0.8523 & 0.8029 & 0.8820 & 0.7768 & 0.8257 \\
& & 150 & 0.8151 & 0.9001 & 0.7589 & 0.8510 & 0.8163 & 0.8953 & 0.7946 & 0.8437 \\
& & 200 & 0.8285 & 0.9101 & 0.7946 & 0.8695 & 0.8285 & 0.9058 & 0.8304 & 0.8615 \\
& & 250 & 0.8229 & 0.9111 & 0.7946 & 0.8665 & 0.8263 & 0.9101 & 0.8036 & 0.8654 \\ \hline
\multirow{12}{*}{NN}&3 & 100 & 0.8296 & 0.9137 & 0.8036 & 0.8691 & 0.8263 & 0.9061 & 0.7857 & 0.8679 \\
&3 & 150 & 0.7996 & 0.8942 & 0.7589 & 0.8437 & 0.8096 & 0.8975 & 0.7679 & 0.8601 \\
&3 & 200 & 0.8408 & 0.9224 & 0.7857 & 0.8735 & 0.8318 & 0.9156 & 0.7946 & 0.8777 \\
\rowcolor{gray}&3 & 250 & 0.8374 & 0.9235 & 0.7768 & 0.8690 & 0.8374 & 0.9205 & 0.8125 & \textbf{0.8792} \\ 
&5 & 100 & 0.8285 & 0.9057 & 0.7857 & 0.8624 & 0.8007 & 0.8854 & 0.7679 & 0.8427 \\
&5 & 150 & 0.8151 & 0.8906 & 0.7679 & 0.8367 & 0.8051 & 0.8830 & 0.7768 & 0.8481 \\
&5 & 200 & 0.8452 & 0.9228 & 0.8036 & 0.8633 & 0.8352 & 0.9172 & 0.8214 & 0.8774 \\
&5 & 250 & 0.8530 & 0.9228 &  0.8036 & 0.8565 & 0.8352 & 0.9159 & 0.8125 & 0.8736 \\ 
&10 & 100 & 0.8218 & 0.9083 & 0.7857 & 0.8632 & 0.8196 & 0.8976 & 0.7679 & 0.8593 \\
&10 & 150 & 0.8140 & 0.8909 & 0.7768 & 0.8461 & 0.8040 & 0.8898 & 0.7768 & 0.8601 \\
&10 & 200 & 0.8463 & 0.9285 & 0.7946 & 0.8724 & 0.8474 & 0.9234 & 0.8125 & 0.8760 \\
&10 & 250 & 0.8419 & 0.9268 & 0.7768 & 0.8709 & 0.8441 & 0.9242 & 0.8125 & 0.8728 \\ \hline
\end{tabular}

\end{table}

Table \ref{tab:modres} presents the results of our grid search in terms of accuracy and AUC, obtained on both the training and validation sets, on the two datasets (original and feature selected) discussed earlier in this section. Note that ``Tr'' stands for ``training set'' and ``Val'' for ``validation set''. We varied the number of training epochs from 100 to 250 for all models. We also varied the number of hidden nodes in our neural network from 3 to 10. Note that logistic regression can be viewed as a neural network with no hidden nodes and layers.  The Adam optimizer was used for training all models with an initial learning rate set to $0.01$ \cite{kingma15}. The best performing model $f^{*}$ is obtained using a neural network with three hidden nodes (AUC: 0.8792) on the ``feature selection'' dataset. Therefore we adopt this model and dataset for the remainder of our experiments.

\begin{figure}[ht]
\begin{subfigure}{.5\textwidth}
 \centering
 %  include second image
  \includegraphics[width=\linewidth]{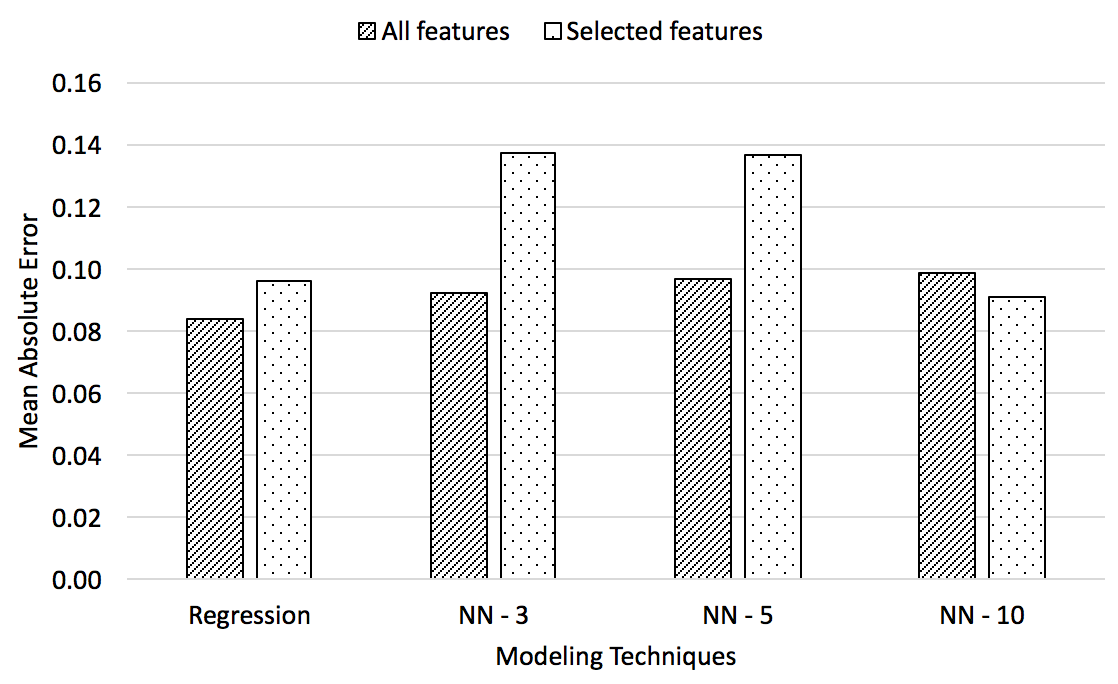}
  \caption{Mean absolute error}
 \label{fig:ife_mae}
\end{subfigure}
\begin{subfigure}{.5\textwidth}
  \centering
  %include first image
  \includegraphics[width=\linewidth]{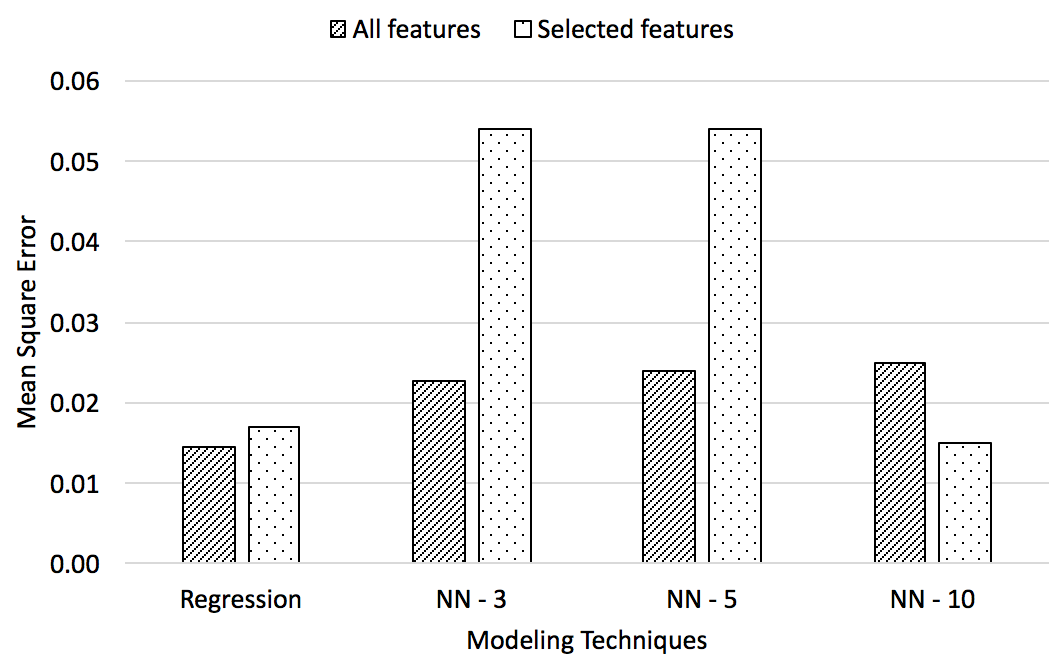}
  \caption{Mean square error}
  \label{fig:ife_mse}
\end{subfigure}
\caption{Model performance of indirect feature estimators on both the ``original'' and ``feature selection'' datasets on a randomly held out validation set. NN-3, NN-5 and NN-10 are the neural network with 3, 5 and 10 hidden nodes, respectively}
 \label{fig:ife}
\end{figure}

Next, we execute a grid search to find the optimal IFE function $H^{*}$. Recall that the IFE takes as input $\bx_U$ and $\bx_D$ and provides estimates for $\bx_I$ and will be used during the recommendation procedure. Figure \ref{fig:ife} illustrates the performance of the model in the validation dataset with 250 epochs. We explored epochs ranging from 100 to 350 and found that 250 epochs produces the best results. Figures \ref{fig:ife_mae} and \ref{fig:ife_mse} show Mean Absolute Error (MAE) and Mean Square Error (MSE), respectively. We utilized MSE as a performance measure to select the best model. The results show that a neural network with ten hidden nodes produces the best model with an MSE of 0.015.

Table \ref{tab:final-models} lists the selected models that will be used in subsequent experiments.

\begin{table}[ht!]
    \centering
    \caption{Selected modeling technique (star indicates the best selected model)}
    \begin{tabular}{cl}
    \hline
        Functions &  Selected Model\\
        \hline
        $f^{*}$ & Neural network with three hidden nodes and sigmoid output activation\\
        $H^{*}$ & Neural network with ten hidden nodes with no output activation\\
        \hline
    \end{tabular}
    
    \label{tab:final-models}
\end{table}

\subsection{Human in the Loop Recommendations: Probability Improvement} \label{sec:ic-prob-improv}

Using the optimal $f^*(\cdot)$ and $H^{*}(\cdot)$, discovered in the preceding subsection, we apply our human in the loop inverse classification formulation, discussed in Section \ref{sec:optimizationModel}, to the test set. The experiments are performed by varying the budget $b$ from 0.1 to 1 with an increment of 0.1. In these experiments the IV fluid values, $\bx_D$, specified by each physician are cumulatively allowed to be changed by only an amount $b$, according to our formulation in Section \ref{sec:optimizationModel}. Therefore, larger values of $b$ will allow larger changes to be made to the physicians recommendation.

\begin{figure}[ht!]
    \begin{subfigure}[t]{0.48\textwidth}
        \centering
        \includegraphics[scale=0.39]{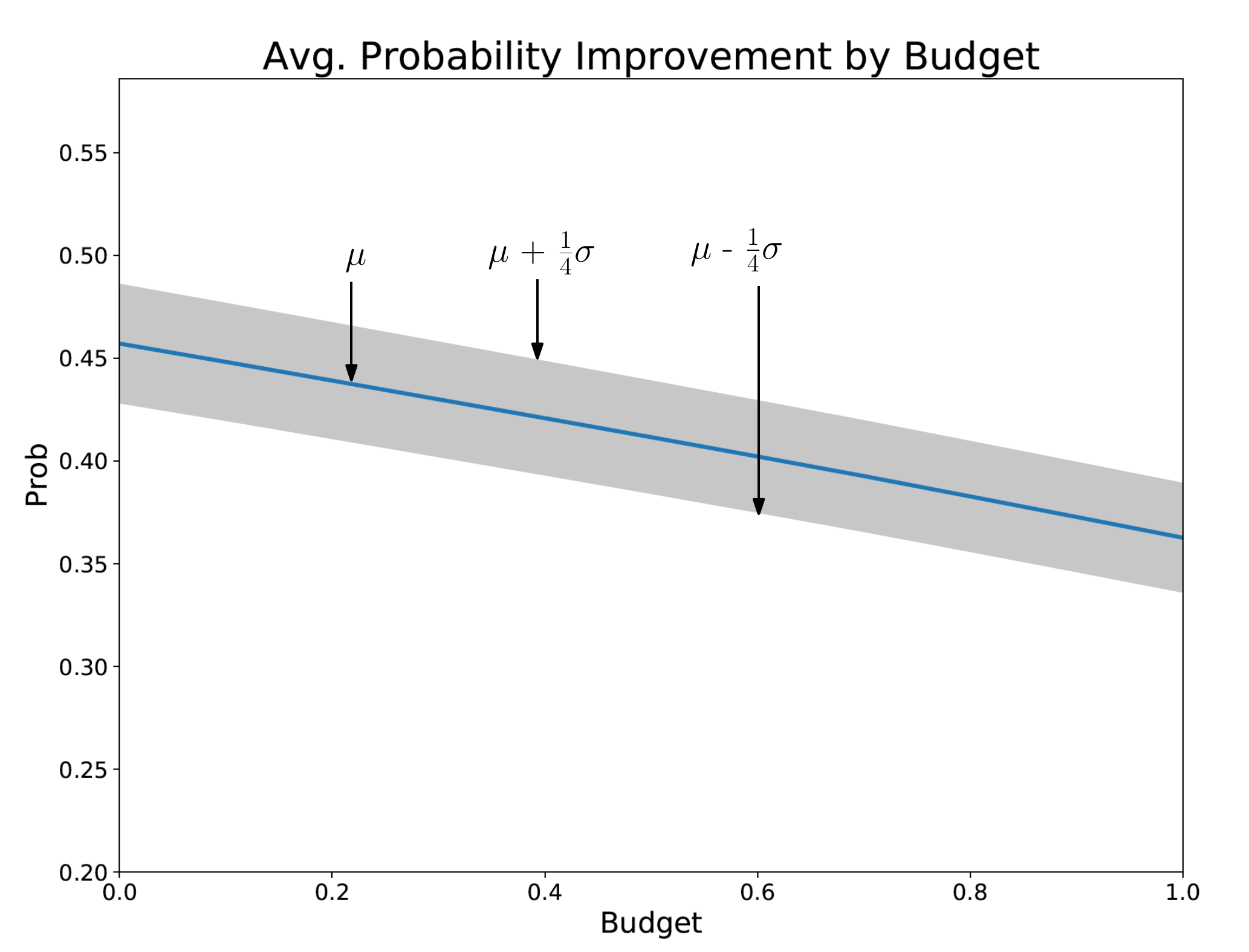}
        \caption{Average probabilities of mortality(along with quarter of one standard deviation) vs. budget.}
         \label{fig:ic-prob-reg}
    \end{subfigure}\hfill
    \begin{subfigure}[t]{0.49\textwidth}
        \centering
        \includegraphics[scale=0.39]{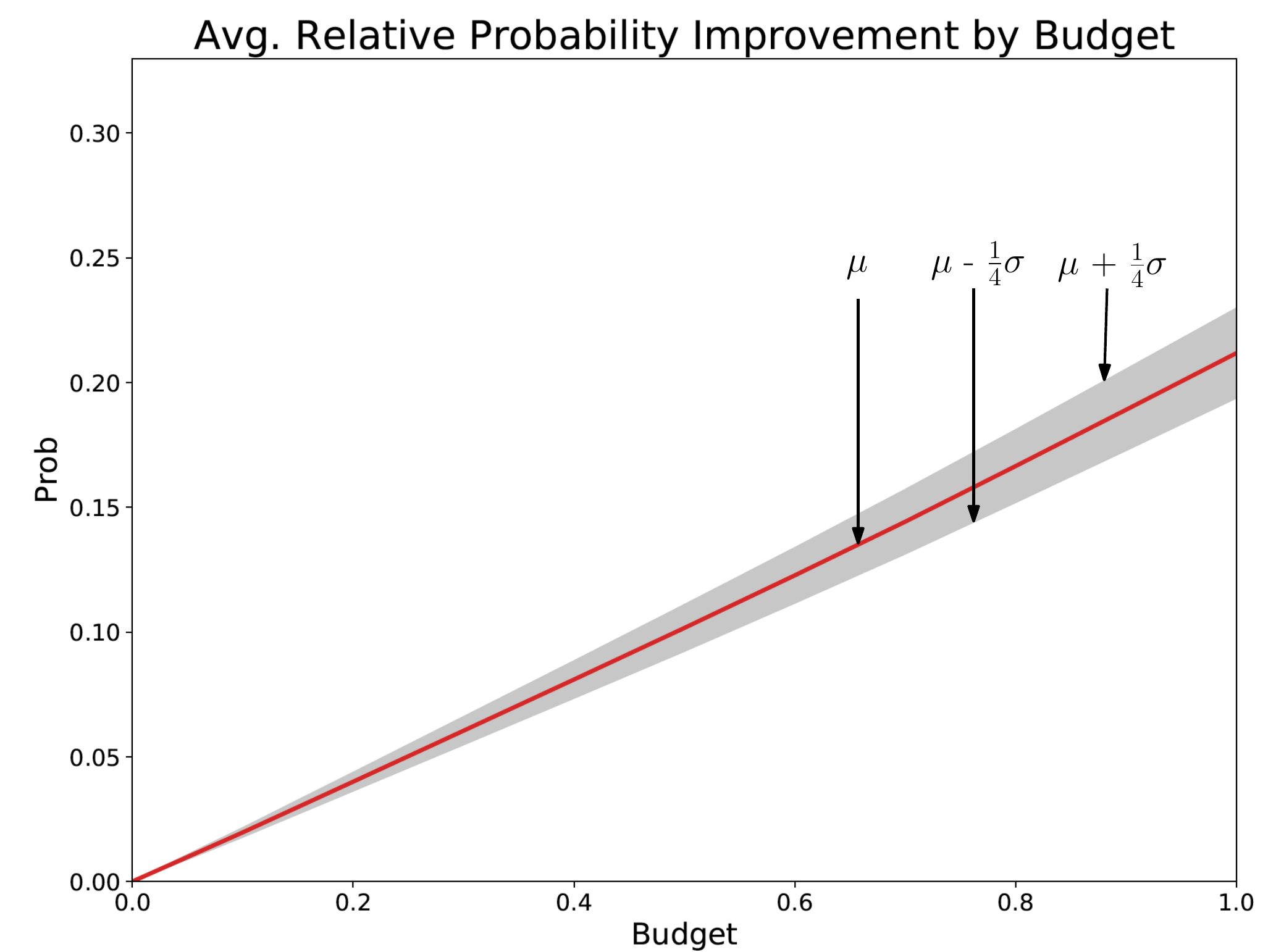}
        \caption{Average relative probabilities of mortality (along with quarter of one standard deviation) vs. budget.}
         \label{fig:ic-prob-rel}
    \end{subfigure}
   \caption{The average change in the probability of mortality over varying budgets.} \label{fig:ic-prob}
   \label{fig:ic-prob}
\end{figure}

Our method derives optimal, patient-specific IV fluid recommendations along with an estimated risk (probability) of mortality (\textit{objective function}) at the varying budget levels mentioned. Figure \ref{fig:ic-prob} shows the average results across these different budget levels. Figure \ref{fig:ic-prob-reg} shows the average ($\mu$) probability of mortality (i.e.,~death due to sepsis; y-axis) at each budget level (x-axis), as indicated by the \textcolor{blue}{blue} line; the \textcolor{gray}{gray} shading indicates one quarter of one standard deviation above and below the average ($\mu \pm \frac{1}{4}\sigma$). As expected, with increase in $b$, the probability of mortality is further reduced (on average). The average probability of mortality with no adjustment in $\bx_D$ is 0.46, while the average probability of mortality with $\bx_D$ allowed to be adjusted to the maximum (i.e.,~$b=1.0$) is 0.37.

Figure \ref{fig:ic-prob-rel} shows relative probability improvement (y-axis) at each budget setting (x-axis), indicated by the the \textcolor{red}{red} line; the \textcolor{gray}{gray} shading indicates one quarter of one standard deviation above and below the average. With increase in $b$, the probability of mortality is further reduced (on average). We can observe that the average relative improvement in mortality, provided limited budget $b=1.0$, is about 22\%, which is significantly better than the 1.8-3.6\% improvement devised by \cite{raghu2017} . Therefore, the proposed model can significantly improve the chances of survival by adjusting the infusion of IV fluids to the optimal value. 

\subsection{Human in the Loop Recommendations: Robustness}\label{sec:ic-robustness}

In Section \ref{sec:ic-prob-improv}, we applied our proposed methodology to our dataset and showed the benefit in terms of average probability improvement. The proposed model uses a physician's recommendations to determine optimal, patient-specific IV fluid dosing. However, we also wanted to investigate the robustness of our model in the absence of any physician input. We refer to the scenarios where the physician's recommendations are incorporated as \textit{human-in-the-loop} initialization, while the scenarios with no physician input are referred to as \textit{random} initialization. We are use the term \textit{initialization} because the input $\bx_D$ values represent the starting place for the optimization procedure. Therefore, the specified $\bx_D$ values likely have an impact on the recommendation and, consequently, the amount of probability improvement that can be extended to each patient (test instance). Therefore, we compare the results obtained using physician inputs to those obtained using random inputs. For each test instance, we randomly initialized each of the $\bx_D$ features to values in the range of $[0,0.1]$. This range of values was selected to reflect a \textit{cautious} initialization (i.e., small, rather than large values).

\begin{figure}[!htp]
    \centering
    \begin{subfigure}[htp]{0.48\textwidth}
        \centering
        \includegraphics[scale=0.38]{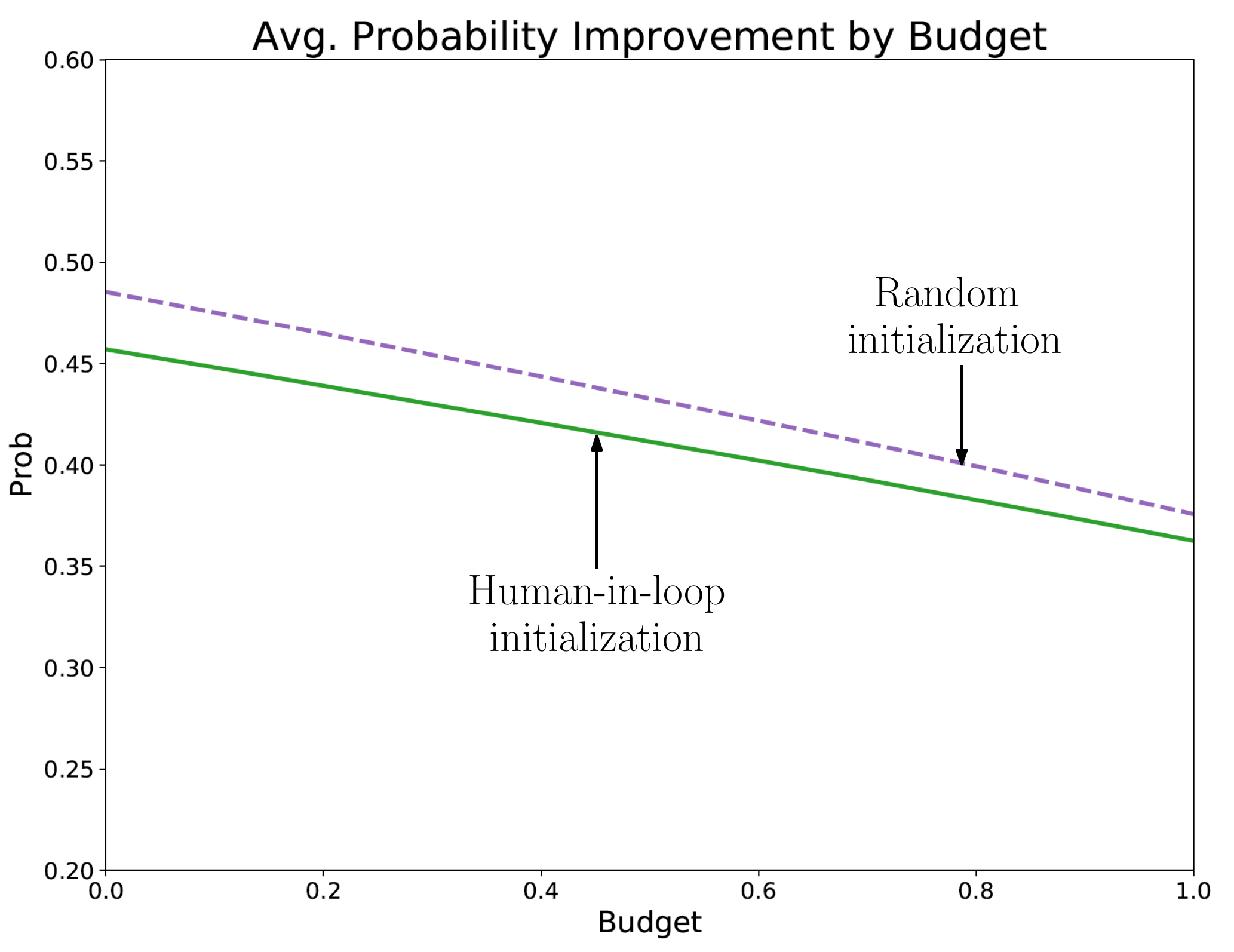}
        \caption{Comparison of average probabilities of mortality considering inverse classifier with human-in-the-loop initialization and with the random initialization.}
        \label{fig:ic-prob-reg-rndm}
    \end{subfigure}\hfill
    \begin{subfigure}[htp]{0.49\textwidth}
        \centering
        \includegraphics[scale=0.39]{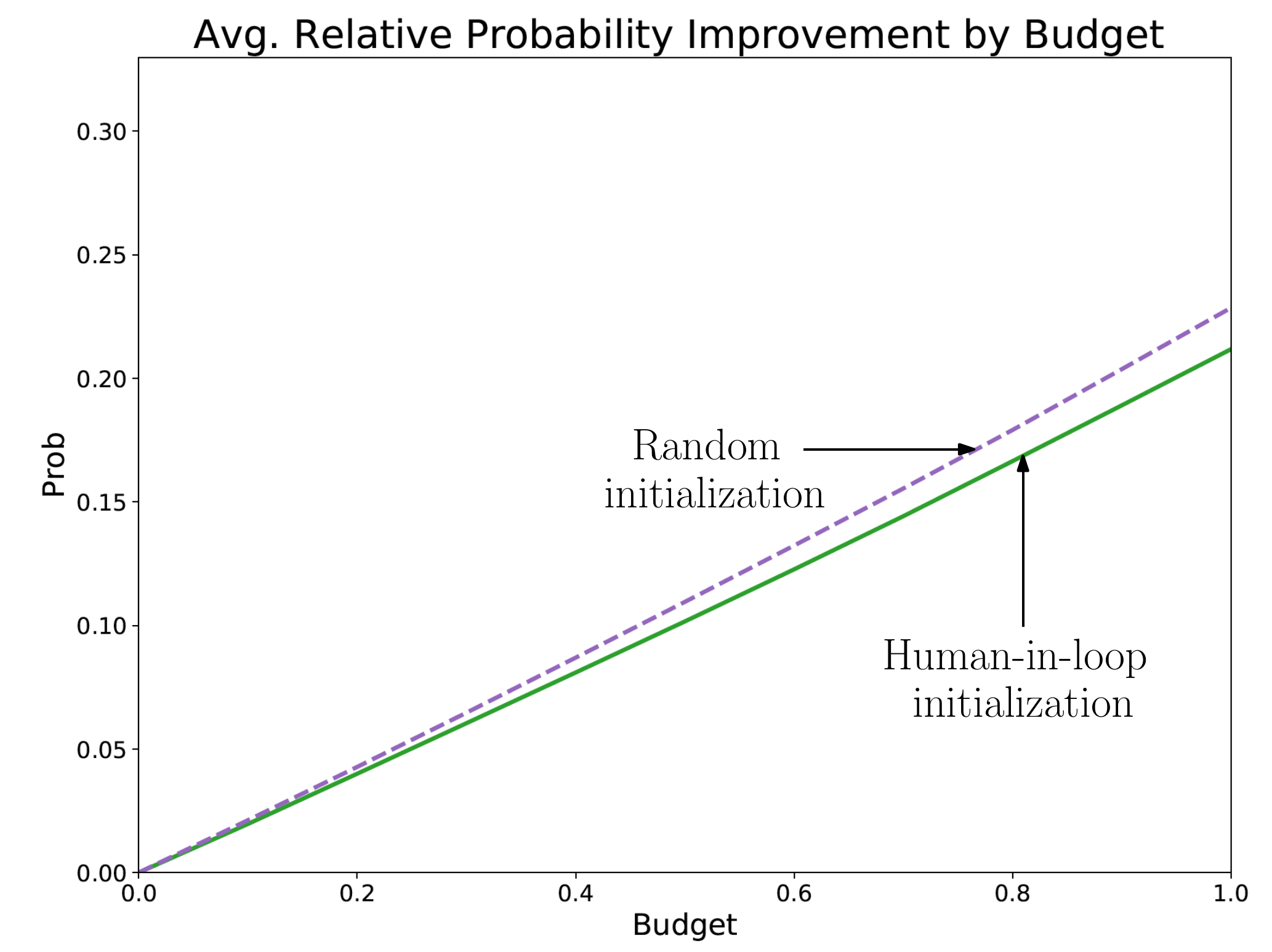}
        \caption{Comparison of relative probability improvement in mortality considering inverse classifier with human-in-the-loop initialization and with the random initialization.}
         \label{fig:ic-prob-rel-rndm}
    \end{subfigure}
    \caption{Human-in-the-loop vs. random initialization results. \label{fig:robust}}
\end{figure}

Figure \ref{fig:robust} illustrates the results obtained using human-in-the-loop initialization and random initialization. Similar to the results presented in the preceding subsection, we show both actual and relative probability improvement at varying budget levels. Figure \ref{fig:ic-prob-reg-rndm} show the average probability improvement results. The x-axis represents the budget and y-axis represents average probability of mortality. The \textcolor{green}{green} line represents human-in-the-loop initialization and the \textcolor{purple}{purple} line represents random initialization. Clearly, integrating a ``human into the loop'' produces reduced mortality results, as compared to the random initialization result. The results demonstrate the importance of the ``human in the loop'' component of our formulation. Nevertheless, the results also show that poor (i.e., random) initializations can still be turned into recommendations that provide comparable benefits (in terms of probability improvement).

Similarly, Figure \ref{fig:ic-prob-rel-rndm} shows relative probability improvement across varying budget levels using both types of initialization.  Again, the \textcolor{green}{green} line represents human-in-the-loop initialization and the \textcolor{purple}{purple} line represents random initialization. Here, we make the same observations that we do for Figure \ref{fig:ic-prob-reg-rndm}. However, we can also see that as the budget is increased, the results obtained from random initialization tends closer to those obtained from physician initialization. This observation further shows that, provided a sufficiently large budget, random initialization can come close to providing the same probabilistic improvement.

\subsection{Human in the Loop Recommendations: Average Recommendations}

In this section, we present the average recommendations made by our human in the loop method. While many scoring criteria \cite{ferreira2001, gupta2018, vincent1996} and tools \cite{gupta2020, henry2015} exist to assess the risk of mortality of septic patients, there are limited studies that comprehensively assess risk, account for physician input, and provide treatment recommendations. Our study and proposed method provides all three of these benefits. Therefore, in this section we examine the recommendations produced by our method. It is impractical, however, to present individual recommendations and, as such, present average recommendations by budget value.

\begin{figure}[!htp]
    %\centering
    %\begin{subfigure}[t]{0.499\textwidth}
       % \centering
      %  \includegraphics[scale=0.24]{results/bud_01.pdf}
     %   \caption{Budget = $0.1$.}
    %\end{subfigure}%
    \centering
    \begin{subfigure}[t]{0.499\textwidth}
        \centering
        \includegraphics[scale=0.24]{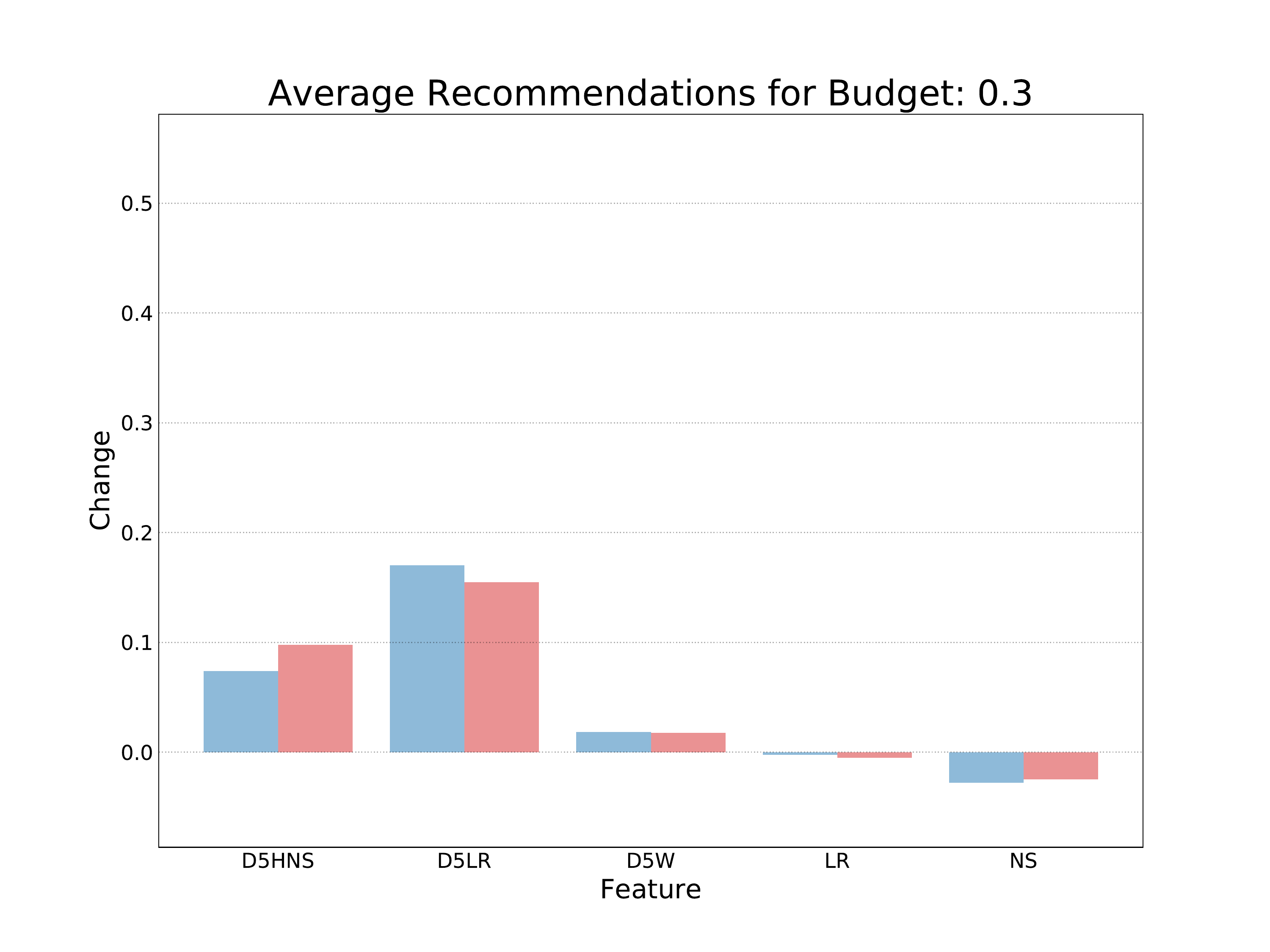}
        \caption{Budget = $0.3$, estimated mortality = 0.44.}
        \label{fig:rec-0.3}
    \end{subfigure}%
    \begin{subfigure}[t]{0.499\textwidth}
        \centering
        \includegraphics[scale=0.24]{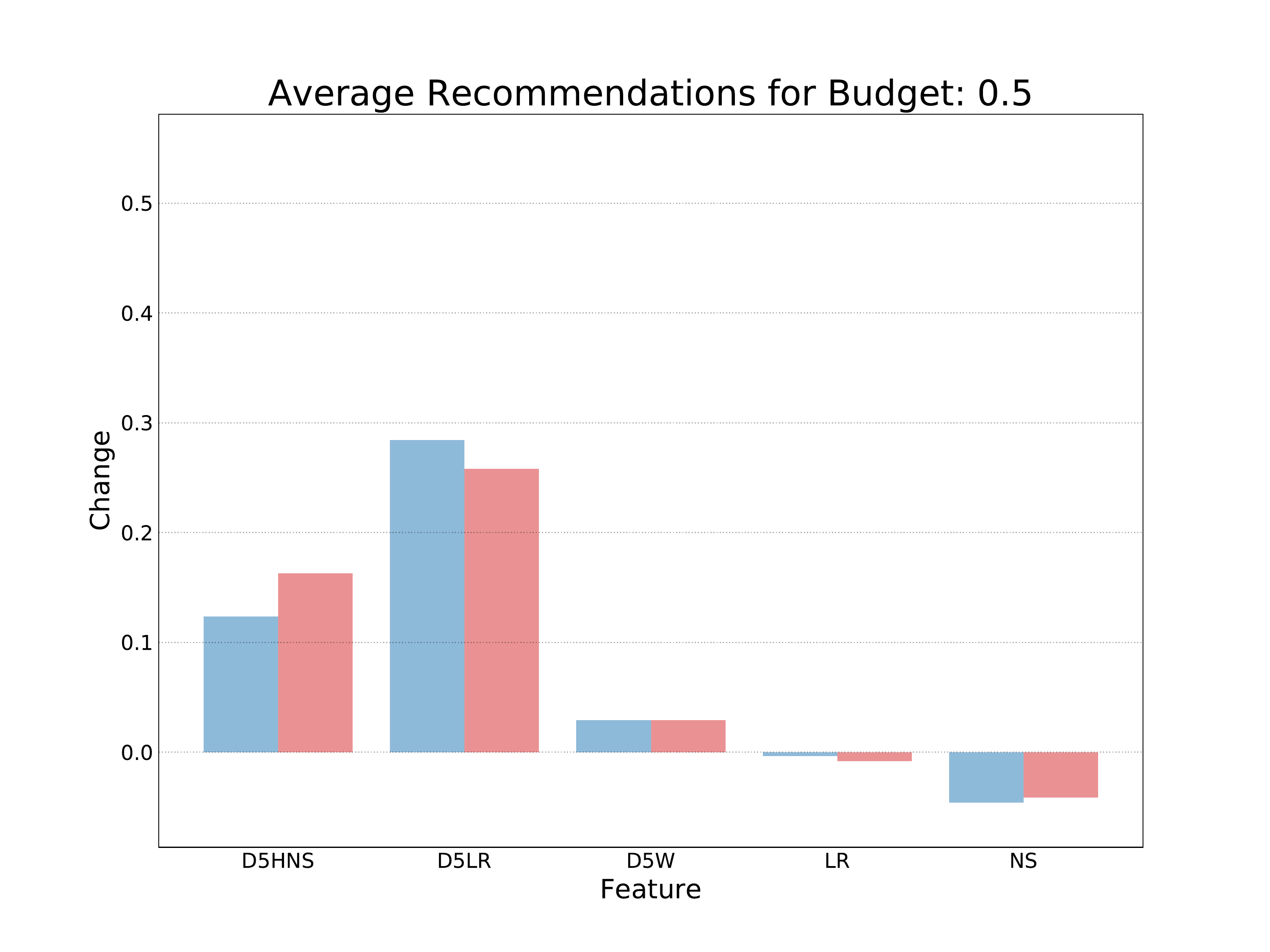}
        \caption{Budget = $0.5$, estimated mortality = 0.43.}
    \end{subfigure}%
    
    \centering
    \begin{subfigure}[t]{0.499\textwidth}
        \centering
        \includegraphics[scale=0.24]{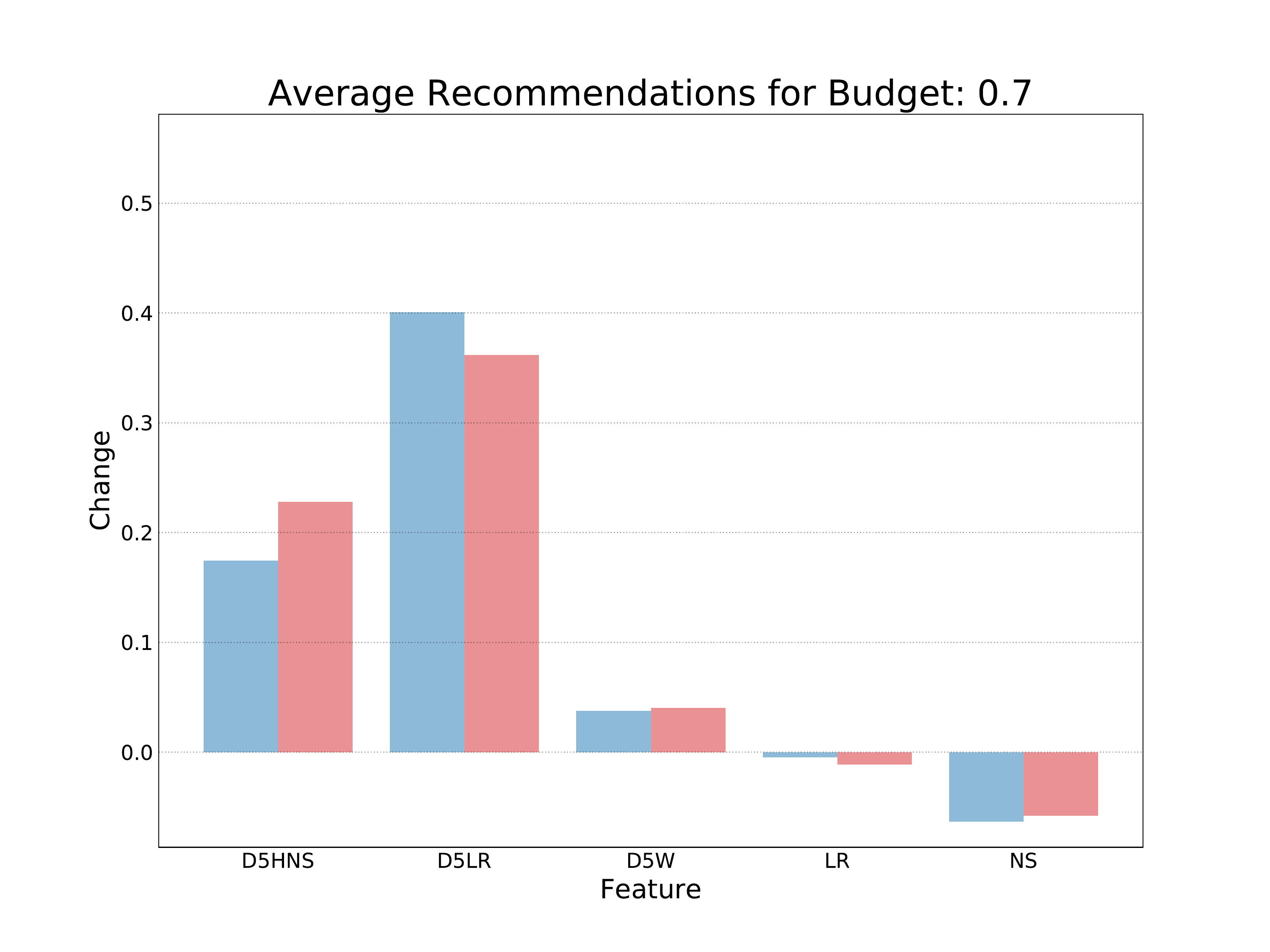}
        \caption{Budget = $0.7$, estimated mortality = 0.40}
    \end{subfigure}%
    \begin{subfigure}[t]{0.499\textwidth}
        \centering
        \includegraphics[scale=0.24]{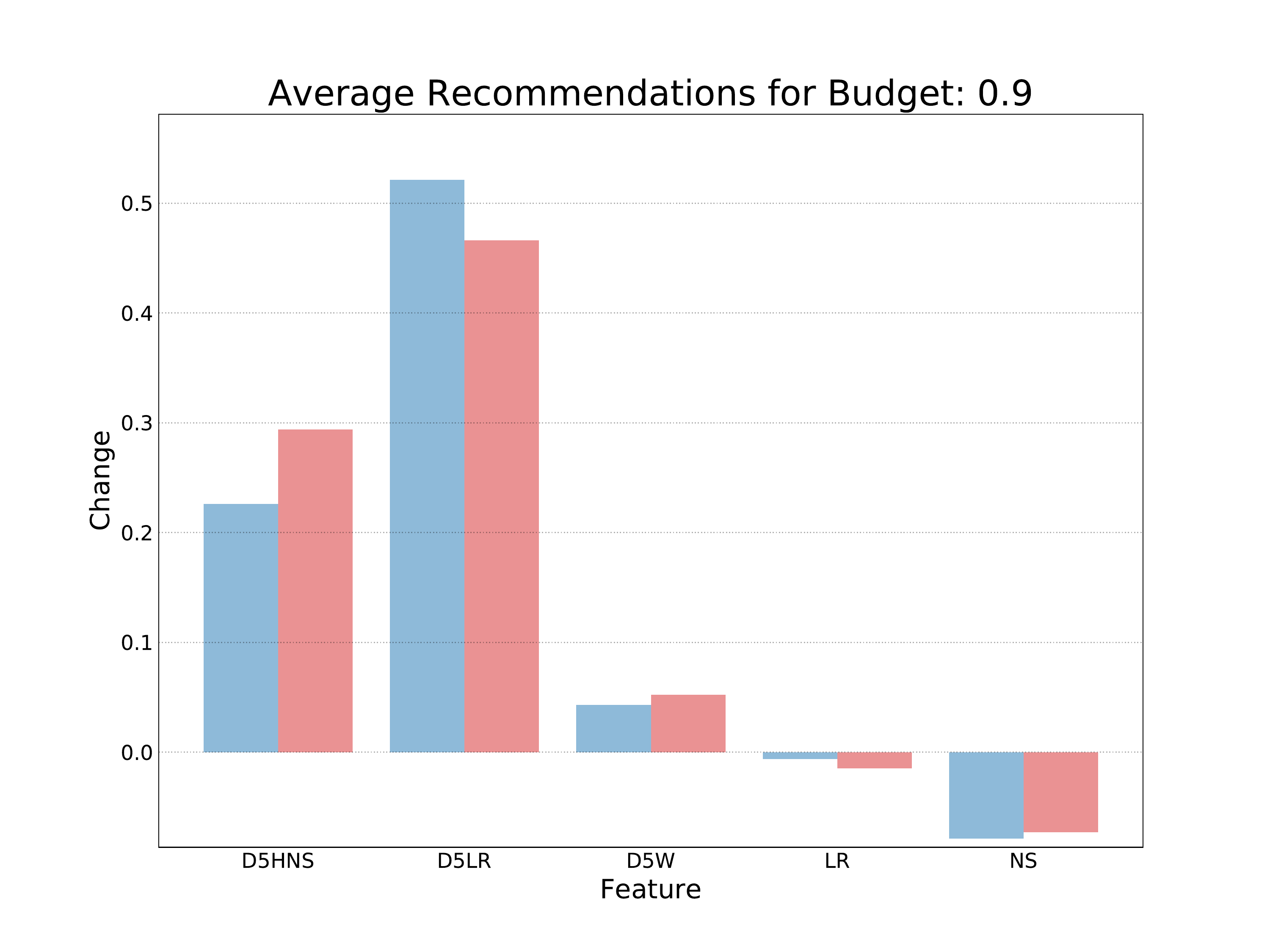}
        \caption{Budget = $0.9$, estimated mortality = 0.38.}
        \label{fig:rec-0.9}
    \end{subfigure}%
    %\centering
    %\begin{subfigure}[t]{0.499\textwidth}
    %    \centering
    %    \includegraphics[scale=0.24]{results/neg_onlybud_09.pdf}
    %    \caption{Negative. Budget = $0.9$.}
    %\end{subfigure}%
    \caption{Average recommendations at budget levels 0.3, 0.5, 0.7, and 0.9, stratified by predicted positive test instances, %(i.e.,~died and survived),
    shown in \textcolor{red}{red}, and predicted negative test instances, shown in \textcolor{blue}{blue}.}
    \label{fig:avg-recs}
\end{figure}

Figure \ref{fig:avg-recs} shows the average recommendation results obtained from applying our method. Each of the Figures \ref{fig:rec-0.3} to \ref{fig:rec-0.9} represent a different budget level (0.3, 0.5, 0.7 and 0.9). The x-axis shows each IV fluid. The y-axis represents the average recommended change to a physicians input. Therefore, deviation from a value of zero represents a recommended change. A positive (negative) value indicates that the physicians recommended IV fluid value should be increased (decreased). The results are further stratified by ``predicted positive'', shown in \textcolor{red}{red}, and ``predicted negative'' shown in \textcolor{blue}{blue}. An instance was predicted as belonging to the positive class, representing ``expiration due to sepsis'', if their predicted probability of mortality was greater than 50\%, and negative otherwise.

Upon inspecting Figures \ref{fig:rec-0.3} through  \ref{fig:rec-0.9} we can clearly see that an increased budget allows greater changes to be made to a physicians suggested IV fluid values. For example, with budget of 0.3, the average suggested change for D5LR for positive and negative predicted cases is 0.15 and 0.17, respectively. While the same for budge of 0.9 is 0.47 and 0.51, respectively. These figures also provide the following insights:
\begin{enumerate}
    \item On average, our model suggests increasing the intake of D5LR, D5HNS and D5W, while recommending that LR and NS be decreased. These observations are in alignment with the literature \cite{huang1995} that suggests that resuscitation using LR is associated with increased renal failure.
    \item Resuscitation using 5\% dextrose in lactate ringer is encouraged among septic patients in ICUs to improve the probability of survival.
    %\item Patients with lower predicted probabilities are recommended to be infused with a bit more D5LR than D5HNS, while patients with higher predicted probabilities are recommended to be infused with a bit more D5HNS than D5LR.
    %\item Patients with lower predicted probabilities are recommended to be infused with a bit more NS than LR, while patients with higher predicted probabilities are recommended to be infused with a bit more LR than NS.
\end{enumerate}

\section{Conclusions}\label{sec:conclusion}

This study proposes a clinical prescriptive model with human in the loop functionality that recommends optimal, individual-specific amounts of IV fluids for the treatment of septic patients in ICUs. The proposed methodology combines constrained optimization and machine learning techniques to arrive at optimal solutions. A key novelty of the proposed clinical model is utilization of a physician's input to derive optimal solutions. 
The efficacy of the method is demonstrated using a real world medical dataset. We further validated the robustness of the proposed approach to show that our method benefits from the human in the loop component, but is also robust to poor input, which is a crucial consideration for new physicians. The results showed, under the limited budget, the optimal solution can improve the average relative probability of survival by 22\%. The proposed method can potentially be embedded in an existing electronic health record system to make life-saving IV fluid recommendations. This model can also be used for training junior physicians to synthesize the appropriate treatment strategy, and prevent user error after the inclusion of additional clinical variables and prospective validation.  An important limitation of the model is the non-inclusion of vasopressors and antibiotics, which are two important classes of drugs used to treat sepsis. 

\section*{Acknowledgement}
The authors would like to thank Dr. Rebekah Child, Ph.D., RN, Associate Professor, California State University - Northridge for initially discussing the medical importance of the problem. 

\begin{comment}
{\color{blue} since our outcome in IFE is not binary, logistic regression is not the appropriate}
{\color{blue}include about indirect feature estimator result. }
{\color{blue} hidden nodes or layers}
Results on the test set:
{\color{blue} we need to discuss about $x_D$ and $F_D$}
\end{comment}

\bibliographystyle{ieeetr}
\bibliography{references}	
\begin{appendix}

\end{appendix}
\end{document}